\documentclass[lettersize,journal]{IEEEtran}
\usepackage{amsmath,amsfonts}
\usepackage{array}
\usepackage[caption=false,font=normalsize,labelfont=sf,textfont=sf]{subfig}
\usepackage{textcomp}
\usepackage{stfloats}
\usepackage{url}
\usepackage{verbatim}
\usepackage{graphicx}
\def\BibTeX{{\rm B\kern-.05em{\sc i\kern-.025em b}\kern-.08em
    T\kern-.1667em\lower.7ex\hbox{E}\kern-.125emX}}
\usepackage{balance}
\usepackage{amssymb}
\usepackage{algpseudocode}
\usepackage{algorithm}
\usepackage{setspace}
\usepackage{threeparttable}
\usepackage{booktabs}
\usepackage{makecell}
\usepackage{xcolor}
\usepackage{ulem}
\usepackage{multirow}

\begin{document}

\definecolor{deepgreen}{RGB}{0,120,0}

\title{ROAP: A Reading-Order and Attention-Prior Pipeline for Optimizing Layout Transformers in Key Information Extraction}
\author{Tingwei Xie, Jinxing He, Yonghong Song$^*$
  % \thanks{Manuscript created October, 2020; This work was developed by the IEEE Publication Technology Department. This work is distributed under the \LaTeX \ Project Public License (LPPL) ( http://www.latex-project.org/ ) version 1.3. A copy of the LPPL, version 1.3, is included in the base \LaTeX \ documentation of all distributions of \LaTeX \ released 2003/12/01 or later. The opinions expressed here are entirely that of the author. No warranty is expressed or implied. User assumes all risk.}
}

% \markboth{Journal of \LaTeX\ Class Files,~Vol.~18, No.~9, September~2020}%
% {How to Use the IEEEtran \LaTeX \ Templates}

\maketitle

\begin{abstract}
  % Modified the first sentence to emphasize the critical limitations rather than just background.
  The efficacy of Multimodal Transformers in visually-rich document understanding (VrDU) is critically constrained by two inherent limitations: the lack of explicit modeling for logical reading order and the interference of visual tokens that dilutes attention on textual semantics.
  To address these challenges, this paper presents ROAP, a lightweight and architecture-agnostic pipeline designed to optimize attention distributions in Layout Transformers without altering their pre-trained backbones.
  The proposed pipeline first employs an Adaptive-XY-Gap (AXG-Tree) to robustly extract hierarchical reading sequences from complex layouts. These sequences are then integrated into the attention mechanism via a Reading-Order-Aware Relative Position Bias (RO-RPB). Furthermore, a Textual-Token Sub-block Attention Prior (TT-Prior) is introduced to adaptively suppress visual noise and enhance fine-grained text-text interactions.
  Extensive experiments on the FUNSD and CORD benchmarks demonstrate that ROAP consistently improves the performance of representative backbones, including LayoutLMv3 and GeoLayoutLM.
  These findings confirm that explicitly modeling reading logic and regulating modality interference are critical for robust document understanding, offering a scalable solution for complex layout analysis. The implementation code will be released at \url{https://github.com/KevinYuLei/ROAP}.
  
  % [Background]
  % [背景] 多模态 Transformer 在视觉富文本文档理解（VrDU）中的有效性受到两个固有局限性的严重制约：缺乏对逻辑阅读顺序的显式建模，以及视觉 token 的干扰稀释了对文本语义的关注。
  % [Objective]
  % [目的] 为解决这些挑战，本文提出了 ROAP，这是一种轻量级且与架构无关的管道，旨在不改变预训练骨干网络的前提下优化 Layout Transformer 的注意力分布。
  % [Methods]
  % [方法] 所提出的框架首先采用自适应 XY 间隙（AXG-Tree）算法从复杂布局中稳健地提取层次化阅读序列。随后，这些序列通过阅读顺序感知相对位置偏置（RO-RPB）集成到注意力机制中。此外，引入了 Textual-Token 子块先验（TT-Prior）以自适应地抑制视觉噪声并增强细粒度的文本-文本交互。
  % [Results]
  % [结果] 在 FUNSD 和 CORD 基准上的大量实验表明，ROAP 持续提升了包括 LayoutLMv3 和 GeoLayoutLM 在内的代表性骨干网络的性能。
  % Notably, in the Relation Extraction task, ROAP significantly boosts recall and logical reasoning capabilities by effectively capturing long-range dependencies.
  % 值得注意的是，在关系抽取任务中，ROAP 通过有效捕捉长距离依赖关系，显著提高了召回率和逻辑推理能力。
  % [Conclusion]
  % [结论] 这些发现证实，显式建模阅读逻辑和调节模态干扰对于稳健的文档理解至关重要，为复杂版面分析提供了一种可扩展的解决方案。
  % [Code]
  % [代码] 本模型的实现代码即将于 https://github.com/KevinYuLei/ROAP 公布。
\end{abstract}


\section{Introduction}
\label{sec:introduction}
\IEEEPARstart{V}{isually}-rich Document Understanding (VrDU) has witnessed a paradigm shift with the advent of multimodal Transformers, which jointly model textual, layout, and visual information to interpret complex documents such as forms, receipts, and invoices.
State-of-the-art models, such as LayoutLMv3~\cite{2022Layoutlmv3} and GeoLayoutLM~\cite{2023GeoLayoutlm}, typically rely on 2D spatial embeddings to capture the geometric arrangement of tokens.
However, spatial proximity in a 2D plane does not always translate to semantic continuity.
In scenarios involving multi-column layouts, rotated text, or dense tables, the lack of an explicit logical reading order (RO) often causes models to misinterpret the sequential flow of information, leading to errors in downstream tasks that require rigorous semantic reasoning.
% ---------------------- 中文翻译 ----------------------
% 随着多模态 Transformer 的出现，视觉富文本文档理解（VrDU）见证了范式的转变，这些模型联合建模文本、版面和视觉信息，以解释表单、收据和发票等复杂文档。
% 最先进的模型，如 LayoutLMv3 \citep{2022Layoutlmv3} 和 GeoLayoutLM \citep{2023GeoLayoutlm}，通常依赖 2D 空间嵌入来捕捉 token 的几何排列。
% 然而，2D 平面中的空间邻近性并不总是转化为语义连续性。
% 在涉及多栏布局、旋转文本或密集表格的场景中，缺乏显式的“逻辑阅读顺序”（RO）往往导致模型误解信息的顺序流，从而导致需要严格语义推理的下游任务出现错误。

Furthermore, the integration of visual modalities introduces a secondary challenge: attention interference.
While visual tokens (e.g., image patches) provide essential contextual cues, they can disproportionately dominate the self-attention mechanism due to their high spatial salience and quantity.
This phenomenon, often referred to as modality competition, dilutes the model's focus on textual semantics, particularly within the textual-token interaction sub-blocks.
Standard attention mechanisms lack structural constraints to mitigate this noise, making it difficult for the model to maintain robust context modeling amidst irrelevant visual signals.
% ---------------------- 中文翻译 ----------------------
% 此外，视觉模态的整合引入了第二个挑战：“注意力干扰”。
% 虽然视觉 token（例如图像 patch）提供了必要的上下文线索，但由于其高空间显著性和数量，它们可能会不成比例地主导自注意力机制。
% 这种现象通常被称为模态竞争，它稀释了模型对文本语义的关注，特别是在 textual-token 交互子块内。
% 标准的注意力机制缺乏结构约束来减轻这种噪声，使得模型难以在无关视觉信号中保持鲁棒的上下文建模。

To address these limitations, we propose a novel pipeline that harmonizes logical reading order with robust attention mechanisms.
We first introduce the Adaptive-XY-Gap Tree (AXG-Tree), a data-driven algorithm that recursively parses document layouts to generate hierarchically consistent reading sequences, robust to skew and irregular spacing.
Building upon this, we present \textbf{ROAP}, a unified optimization pipeline designed to inject this logical order into the Transformer architecture.
% ---------------------- 中文翻译 ----------------------
% 为了解决这些局限性，我们提出了一个新的框架，将逻辑阅读顺序与鲁棒的注意力机制相协调。
% 我们首先介绍了“自适应 XY 间隙树（AXG-Tree）”，这是一种数据驱动的算法，递归地解析文档布局以生成层次一致的阅读序列，对倾斜和不规则间距具有鲁棒性。
% 在此基础上，我们提出了 **ROAP**，这是一个统一的优化管道，旨在将这种逻辑顺序注入到 Transformer 架构中。

\textbf{ROAP} is a lightweight and architecture-agnostic optimization pipeline that augments layout-aware Transformers by explicitly encoding AXG-Tree-derived sequences via \textbf{R}eading-\textbf{O}rder-Aware Relative Position Biases (RO-RPB) and utilizing a Textual-Token sub-block \textbf{A}ttention \textbf{P}rior (TT-Prior) to filter visual noise.
Crucially, it can be seamlessly integrated into any self-attention-based document understanding model, including LayoutLMv3 and GeoLayoutLM, without altering their original architectures.
% ---------------------- 中文翻译 ----------------------
% **ROAP 是一个轻量级且与架构无关的优化管道，它通过阅读顺序感知的相对位置偏置（RO-RPB）显式编码源自 AXG-Tree 的序列，并利用 Textual-Token 子块先验（TT-Prior）来过滤视觉噪声，从而增强布局感知 Transformer。**
% 至关重要的是，**它可以无缝集成到任何基于自注意力的文档理解模型中，包括 LayoutLMv3 和 GeoLayoutLM，而无需更改其原始架构。**

Our contributions are summarized as follows:
\begin{itemize}
    \setlength\itemsep{0em} % 设置列表项之间的间距为0，保持紧凑
    \item[(1)] We propose the Adaptive-XY-Gap Tree (AXG-Tree) for robust reading order extraction in complex layouts.
    % (1) 我们提出了 AXG-Tree 算法，用于复杂布局中的鲁棒阅读顺序提取。

    \item[(2)] We introduce the Reading-Order-Aware Relative Position Bias (RO-RPB) to explicitly encode logical sequences into the self-attention mechanism without disrupting the original spatial perception.
    % (2) 我们引入了阅读顺序感知相对位置偏置（RO-RPB），以便在不破坏原有空间感知的情况下，将逻辑序列显式编码到自注意力机制中。

    \item[(3)] We design the Textual-Token Sub-block Attention Prior (TT-Prior) to adaptively suppress visual modality noise and refine fine-grained text-text interactions.
    % (3) 我们设计了 Textual-Token 子块先验（TT-Prior），以自适应地抑制视觉模态噪声并优化细粒度的文本-文本交互。

    \item[(4)] Extensive experiments on FUNSD and CORD benchmarks demonstrate that ROAP consistently improves performance, achieving significant gains on the Semantic Entity Recognition (SER) and Relation Extraction (RE) tasks.
    % (4) 在 FUNSD 和 CORD 基准上的大量实验表明，ROAP 持续提升了性能，并在语义实体识别（SER）和关系抽取（RE）任务上取得了显著增益。
\end{itemize}

\section{Related Work}
\label{sec:related work}

\subsection{Reading Order Modeling in Document Understanding}

Reading order (RO) information plays a pivotal role in VrDU, as it determines how textual and layout cues are sequentially perceived to form coherent semantics. 
Early layout-aware models often employed heuristic algorithms such as XY-Cut~\cite{2022XYLayoutlm} to infer reading sequences from the geometric arrangements of text boxes. 
Although these rule-based methods demonstrate the feasibility of deriving RO directly from spatial layouts, they are sensitive to irregular alignments, skewed orientations, and multi-column structures, which greatly limit their robustness across diverse document formats.
% 阅读顺序（Reading Order, RO）信息在视觉富文本文档理解（VrDU）中至关重要，因为它决定了文本与版面线索在语义上被感知的逻辑顺序。
% 早期的版面感知模型通常采用启发式算法（如 XY-Cut~\citep{2022XYLayoutlm}），根据文本框的几何排列推断阅读顺序。
% 尽管这类基于规则的方法验证了从空间布局直接提取阅读顺序的可行性，但它们对非规则排版、文本倾斜和多栏结构高度敏感，因而在不同文档格式中鲁棒性较差。

\begin{figure*}[t]
    \centering
    \includegraphics[width=\textwidth]{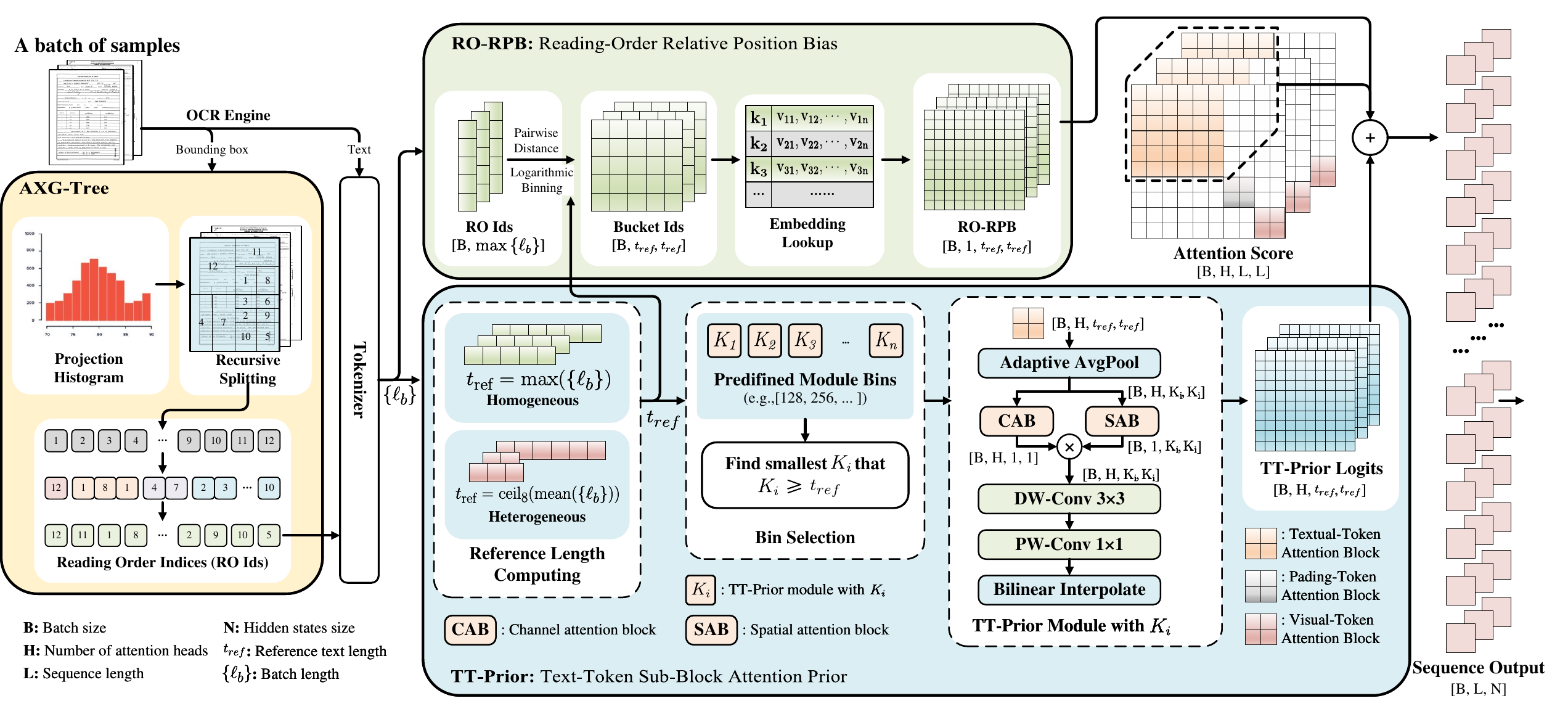}
    \caption{Overall architecture of the proposed ROAP pipline.}
    \label{fig:roap_architecture}
\end{figure*}

Subsequent studies have highlighted that accurate RO modeling can substantially enhance downstream performance.
~\cite{2023-TPP} proposed the Token Path Prediction (TPP) framework to address disordered input issues in OCR-based document understanding. By predicting entity mentions as token paths rather than sequential labels, TPP effectively mitigates the dependence on strictly ordered token sequences. 
However, this approach introduces a dense graph structure among tokens, leading to increased computational complexity and memory cost, and its reliance on grid-level prediction makes integration into pre-trained multimodal frameworks less straightforward. 
Similarly,~\cite{2022-ernie} incorporated layout-aware reading order knowledge into the pre-training objectives of multimodal transformers, achieving performance gains on form and receipt understanding tasks. Nevertheless, this method demands extensive labeled data and high computational resources during pre-training, restricting its scalability and flexibility in fine-tuning stages.
% 随后的研究表明，精确的阅读顺序建模能显著提升下游任务性能。
% \citet{2023-TPP} 提出了 Token Path Prediction (TPP) 框架，通过将实体识别建模为 token 路径预测而非序列标注，有效缓解了 OCR 结果中输入次序紊乱的问题。然而，该方法需要在 token 之间构建稠密图结构，计算与显存开销较大，同时其基于网格的预测形式也使其难以直接与多模态预训练模型兼容。
% 类似地，\citet{ernie-layout} 在多模态 Transformer 预训练阶段引入了阅读顺序预测目标，并在表单和收据理解任务中取得显著性能提升。但该方法依赖大规模标注数据与高计算资源，限制了其可扩展性和微调阶段的灵活性。

Recent work such as~\cite{2024-roor} further formalized reading order prediction as an ordering-relation extraction problem, where models learn pairwise order relations among text entities to reconstruct logical reading paths. 
Although conceptually elegant, these relation-based methods often require detailed supervision and complex annotation schemes, which hinder practical deployment in large-scale scenarios.
% 最新的研究如 \citet{2024-roor} 将阅读顺序预测形式化为关系抽取问题，通过学习文本实体之间的成对顺序关系来重建逻辑阅读路径。
% 尽管该方法在理论上具有优雅的形式，但往往需要复杂的监督信号与精细的关系标注，难以在大规模场景下推广

Based on the above observations, this paper proposes the AXG-Tree, which efficiently infers robust reading-order indices and naturally integrates them into transformer attention through relative positional bias encoding.
% 基于上述现状，本文提出 AXG-Tree 算法，旨在高效推断稳健的阅读顺序索引，并通过相对位置偏置的方式自然融入 Transformer 的注意力机制。

\subsection{Attention Interference Between Textual and Visual Tokens}
Transformer-based architectures have become central to VrDU, yet their self-attention mechanism is easily disrupted by the heterogeneous composition of multimodal token sequences. 
In real scanned documents, textual tokens and visual/layout tokens coexist, and prior studies have shown that these modalities often compete for attention capacity.~\cite{abdallah2024survey} highlight that visual embeddings—though similar in number to textual tokens—can draw substantial attention mass due to their high spatial salience, thereby weakening the model’s ability to preserve fine-grained linguistic relationships.
% 基于 Transformer 的视觉富文本文档理解模型虽然性能强大，但其自注意力机制极易受到多模态 token 组成的干扰。
% 在真实扫描文档中，文本 token 与视觉／版式 token 共同存在，多项研究指出这两类 token 在注意力资源上相互竞争。\citet{abdallah2024survey} 指出，即便视觉 token 的数量与文本 token 大体相近，其高空间显著性仍可能吸引大量注意力权重，从而削弱模型保持精细语言关联的能力。

Similar interference phenomena have been observed in other multimodal Transformer systems.
~\cite{nguyen-etal-2021-skim-attention} show that treating all tokens uniformly allows visually dominant but semantically weak regions to attract disproportionate attention, introducing noise into the reasoning process. 
From an efficiency perspective,~\cite{faststructext2023} demonstrate that long multimodal sequences amplify such interference, as visual tokens accumulate rapidly with increasing image resolution. 
Likewise,~\cite{dockylin2024} report that excessive visual tokens complicate attention routing and degrade semantic consistency, particularly in forms and receipts where fine textual distinctions are essential. 
Beyond efficiency considerations,~\cite{toker-etal-2025-padding} provide mechanistic evidence that Transformer attention is highly sensitive to token heterogeneity, further confirming that non-textual tokens can distort attention patterns even without dominating the sequence length.
% 多模态 Transformer 中类似的干扰现象也屡有报道。
% \citet{nguyen-etal-2021-skim-attention} 表明，若对所有 token 一视同仁，视觉上显著但语义较弱的区域会获得过高注意力，导致推理过程出现噪声。
% \citet{faststructext2023} 强调，随着图像分辨率提升，视觉 token 数量迅速增长，使注意力竞争进一步加剧。
% \citet{dockylin2024} 同样指出，视觉 token 过多会复杂化注意力路由并降低语义一致性，尤其在表单与收据场景中影响更为严重。
% 此外，\citet{toker-etal-2025-padding} 从机制层面证明 Transformer 对 token 异质性高度敏感，进一步印证视觉 token 会扭曲注意力模式。

Collectively, these findings reveal that visual tokens, by virtue of their spatial prominence, often compete with textual tokens for attention and obscure meaningful linguistic interactions. 
This interference is most detrimental within the textual-token (TT) sub-block of the attention matrix, where accurate modeling of intra-text relations is crucial for downstream VrDU tasks. 
% 整体来看，视觉 token 由于其空间显著性，往往与文本 token 争夺注意力资源，从而掩盖文本内部的重要语义交互。
% 这一干扰在注意力矩阵的文本–文本（TT）子块中尤为有害，因为文本关系建模对下游任务至关重要。

Motivated by these observations, we introduce a TT-Prior mechanism that injects structural priors into the TT attention sub-block to suppress visual-token-induced noise and enhance attention focus on semantically relevant textual regions.
% 基于上述认识，本文提出 TT-Prior，通过向 TT 子块注入结构先验以抑制视觉 token 带来的干扰，并强化模型对语义相关文本区域的关注。

\section{Methodology}
\label{sec:methods}

\subsection{Overview of the Proposed ROAP Pipeline}
\label{subsec:overview}
The proposed ROAP pipeline enhances Transformer-based document understanding models by explicitly incorporating reading-order information into the attention mechanism. As shown in Figure~\ref{fig:roap_architecture}, the pipeline consists of two complementary components: (1) reading-order guided relative positional encoding, and (2) a textual-token (TT) sub-block attention prior designed to mitigate noise introduced by padding tokens and visual tokens.

% 中文说明：
% 本段介绍ROAP框架的整体目标：通过在注意力机制中显式引入阅读顺序信息，以增强文档理解模型的能力。并指出ROAP包含两个核心部件：RO相对位置偏置和TT先验。
% 提出的ROAP框架通过显式地将阅读顺序信息合并到注意机制中来增强基于transformer的文档理解模型。如图\ref{fig:roap_architecture}所示，该框架由两个互补组件组成：(1)读取顺序引导的相对位置编码，以及(2)标记-标记（TT）子块先验，旨在减轻填充标记和视觉标记引入的噪声。

To obtain reliable reading-order signals, ROAP first applies the Adaptive-XY-Gap Tree (AXG-Tree) to the OCR-detected text boxes. This algorithm organizes text regions into a hierarchical structure and generates a deterministic reading-order index for each token, capturing the document’s natural spatial reading flow without requiring manual annotations. After tokenization, the reading-order indices are aligned with the token sequence and padded to a uniform length within each batch.

% 中文说明：
% 本段描述AXG-Tree如何为每个文本token生成RO ID，并通过tokenization对齐与padding，使得RO ID可以用于Transformer结构。
% 为了获得可靠的读序信号，ROAP首先对ocr检测到的文本框应用Adaptive-XY-Gap （AXG-Tree）算法。该算法将文本区域组织到层次结构中，并为每个标记生成确定的阅读顺序索引，在不需要手动注释的情况下捕获文档的自然空间阅读流。在标记化之后，读取顺序索引与标记序列对齐，并在每个批处理中填充为统一的长度。

The reading-order indices are then used to compute a relative positional bias (RO-RPB) that encodes pairwise ordering relations among text tokens. This bias is injected into the attention logits before softmax, enabling the model to favor reading-order-consistent token interactions.

% 中文说明：
% 本段描述RO相对位置偏置（RO-RPB），以及它如何被注入注意力logits以提升阅读顺序敏感性，同时强调与任意Transformer主干的兼容性。
% 使用读取顺序索引来计算相对位置偏差（RO-RPB），该偏差对文本标记之间的成对排序关系进行编码。这种偏差在softmax之前被注入到注意逻辑中，使模型更倾向于阅读顺序一致的令牌交互。

Complementing the reading-order alignment, ROAP incorporates a Textual-Token Sub-block Attention Prior (TT-Prior) mechanism to mitigate the modality interference inherent in multimodal architectures. 
In standard self-attention, the visual tokens---often abundant and spatially distinct---tend to dominate the attention distribution, thereby diluting the semantic interactions within the textual modality. 
The attention sub-block corresponding to the text-text region is thus susceptible to being overwhelmed by visual noise. 
ROAP addresses this by computing a reference text length for each batch and dynamically routing the attention computation to a matched TT-Prior module. 
The selected prior injects structured biases specifically into the textual-token sub-block, effectively suppressing the interference of visual tokens and reinforcing the semantic coherence among textual elements.
% 为了配合阅读顺序的对齐，ROAP 引入了“文本-Token 子块先验（TT-Prior）”机制，以减轻多模态架构中固有的模态干扰。
% 在标准的自注意力机制中，视觉 Token——通常数量众多且空间特征显著——倾向于主导注意力分布，从而稀释了文本模态内部的语义交互。
% 因此，对应于文本-文本区域的注意力子块很容易被视觉噪声所淹没。
% ROAP 通过为每个批次计算参考文本长度，并将注意力计算动态路由到匹配的 TT-Prior 模块来解决这个问题。
% 所选的先验将结构化偏置专门注入到文本-文本子块中，有效地抑制了视觉 Token 的干扰，并增强了文本元素之间的语义连贯性。

Our ROAP pipeline provides a unified and lightweight enhancement that can be seamlessly integrated into existing document understanding models such as LayoutLMv3, GeoLayoutLM. 
By jointly leveraging global reading-order structure and local sub-block attention priors, ROAP improves both the faithfulness and robustness of Transformer attention, leading to consistent performance gains across datasets and tasks.
% 我们提出的ROAP框架提供了一个统一的轻量级增强，可以无缝地集成到现有的文档理解模型（如LayoutLMv3和GeoLayoutLM）中。
% 通过联合利用全局读取顺序结构和本地子块先验，ROAP提高了Transformer注意的可靠性和健壮性，从而在数据集和任务之间获得一致的性能增益。

\subsection{Adaptive-XY-Gap Tree for Robust Reading Order Extraction}
\label{subsec:axg-tree}

Accurate reading order extraction is fundamental for document understanding, as the sequence of textual elements often determines their semantic relationships. 
Traditional top-down heuristics such as XY-Cut or recursive projection segmentation are highly sensitive to local skew, irregular layouts, and uneven text spacing, which frequently occur in scanned or handwritten documents. 
To address these limitations, we propose an Adaptive-XY-Gap Tree (AXG-Tree) --a lightweight, data-driven hierarchical clustering method that dynamically segments text regions along both axes to construct a robust reading order tree.
% ---------------------- 中文翻译 ----------------------
% 准确的阅读顺序提取是文档理解的基础，因为文本元素的顺序往往决定了它们的语义关系。
% 传统的自顶向下启发式算法（如 XY-Cut 或递归投影分割）对局部倾斜、不规则布局和不均匀的文本间距高度敏感，而这些情况经常出现在扫描或手写文档中。
% 为了解决这些局限性，我们提出了自适应 XY 间隙树（AXG-Tree）算法——一种轻量级、数据驱动的层次聚类方法，它沿两个轴动态分割文本区域以构建稳健的阅读顺序树。

% 第二段：预处理（倾斜校正）+ 核心递归逻辑 + 投影直方图公式
% 重点：解释算法如何通过投影直方图来量化布局结构。
To initiate the process, a global skew correction is applied to align the text boxes horizontally, minimizing the noise caused by document rotation.
The core of AXG-Tree involves a recursive splitting strategy.
At each recursion step with a subset of boxes, we compute a projection histogram $\mathbf{h}$ along the scanning axis (e.g., the Y-axis for row segmentation).
The value at the $k$-th bin, denoted as $\mathbf{h}[k]$, represents the density of text elements projected onto that interval:
\begin{equation}
    \mathbf{h}[k] = \sum_{b \in \mathcal{B}_{sub}} \mathbb{I}(b \cap \text{bin}_k \neq \emptyset),
\end{equation}
where $\mathbb{I}(\cdot)$ is the indicator function, and $\mathcal{B}_{sub}$ is the current subset of boxes.
% ---------------------- 中文翻译 ----------------------
% 为了启动该过程，首先应用全局倾斜校正将文本框水平对齐，以最大程度地减少由文档旋转引起的噪声。
% AXG-Tree 的核心涉及一种递归分割策略。
% 在对文本框子集进行的每一步递归中，我们计算沿扫描轴（例如，用于行分割的 Y 轴）的投影直方图 $\mathbf{h}$。
% 第 $k$ 个 bin 的值（记为 $\mathbf{h}[k]$）表示投影到该区间上的文本元素的密度（公式如上，其中 $\mathbb{I}$ 为指示函数）。

\begin{algorithm}[t]
\caption{Adaptive-XY-Gap Tree (AXG-Tree)}
\label{alg:axgtree}
\begin{algorithmic}[1]
% 定义输入输出格式
\Require Set of text boxes $\mathcal{B} = \{b_1, \dots, b_N\}$, image size $(W, H)$, hyperparameters $\alpha, \beta, \gamma$.
\Ensure Ordered sequence of indices $\mathcal{O}$.

\State \textbf{Preprocessing:} Correct skew of $\mathcal{B}$ if necessary.
\State Initialize index list $\mathcal{I} \leftarrow [1, \dots, N]$.

% 定义递归函数
\Function{Recurse}{$\mathcal{I}_{sub}$, axis}
    \If{$|\mathcal{I}_{sub}| \le 1$}
        \State \Return $\mathcal{I}_{sub}$
    \EndIf
    
    \State Get boxes $\mathcal{S} \leftarrow \{b_i \mid i \in \mathcal{I}_{sub}\}$.
    \State Compute projection histogram $\mathbf{h}$ along \textit{axis}.
    \State Compute median $m$ and IQR of box sizes along \textit{axis}.
    
    \State \textit{// Calculate Thresholds consistent with Eq. (1) \& (2)}
    \State $gap_{min} \leftarrow \alpha \cdot m + \beta \cdot \text{IQR}$
    \State $\tau \leftarrow \gamma \cdot \text{median}(\mathbf{h})$
    
    \State \textbf{Find Valleys:} Identify segments where $\mathbf{h}[k] < \tau$ with width $\ge gap_{min}$.
    \State Split $\mathcal{S}$ into groups $\{G_1, \dots, G_k\}$ based on valleys.
    
    \If{No valid split found}
        \If{\textit{axis} is Y}
            \State \Return \Call{Recurse}{$\mathcal{I}_{sub}$, X} \Comment{Try other axis}
        \Else
            \State \Return \Call{AGS\_Sort}{$\mathcal{I}_{sub}$} \Comment{Fallback}
        \EndIf
    \Else
        \State Sort groups $\{G_j\}$ by coordinate.
        \State $\mathcal{O}_{local} \leftarrow []$
        \For{$j = 1$ to $k$}
            \State $\mathcal{O}_{local}$.append(\Call{Recurse}{$G_j$, opposite \textit{axis}})
        \EndFor
        \State \Return $\mathcal{O}_{local}$
    \EndIf
\EndFunction

\State $\mathcal{O} \leftarrow$ \Call{Recurse}{$\mathcal{I}$, Y} \Comment{Start with vertical Y-axis split}
\State \textbf{Post-process:} Correct missing or duplicate indices if any.
\State \Return $\mathcal{O}$
\end{algorithmic}
\end{algorithm}

% 第三段：自适应阈值公式 (Gap 和 Tau)
% 重点：引入伪代码中用到的两个核心阈值 gap_min 和 tau，并解释其超参数含义。
Instead of using rigid geometric rules, AXG-Tree determines valid separators based on statistical properties.
We define two adaptive thresholds: the minimum gap width $gap_{min}$ and the valley density threshold $\tau$.
The $gap_{min}$ is calculated from the box dimensions perpendicular to the cut (e.g., box heights when splitting rows):
\begin{equation}
    gap_{min} = \alpha \cdot m + \beta \cdot \text{IQR},
\end{equation}
where $m$ and $\text{IQR}$ denote the median and interquartile range of the box sizes.
$\alpha$ and $\beta$ are hyperparameters regulating the baseline sensitivity and tolerance to size dispersion, respectively.
Simultaneously, to distinguish meaningful structural gaps from intra-line spacing, we define the density threshold $\tau$ as:
\begin{equation}
    \tau = \gamma \cdot \text{median}(\mathbf{h}),
\end{equation}
where $\gamma \in (0, 1]$ acts as a noise-tolerance coefficient.
% ---------------------- 中文翻译 ----------------------
% AXG-Tree 不使用僵化的几何规则，而是根据统计特性确定有效的分割符。
% 我们定义了两个自适应阈值：最小间隙宽度 $gap_{min}$ 和谷值密度阈值 $\tau$。
% $gap_{min}$ 是根据垂直于切割方向的文本框尺寸（例如，分割行时使用文本框高度）计算得出的（公式如上，m 和 IQR 分别为中位数和四分位距）。
% $\alpha$ 和 $\beta$ 分别是调节基准灵敏度和尺寸离散度容忍度的超参数。
% 同时，为了区分有意义的结构性间隙和行内间距，我们将密度阈值 $\tau$ 定义为（公式如上），其中 $\gamma$ 充当噪声容忍系数。

% 第四段：Valley检测逻辑 + 伪代码对应 + 兜底策略
% 重点：将文字描述与伪代码中的 "\mathbf{h}[k] < \tau" 和 "width \ge gap_{min}" 对应起来，并说明递归终止条件和 AGS 兜底。
Based on these metrics, the algorithm identifies "valleys"---potential split points---by locating continuous segments where the projection density is sufficiently low ($\mathbf{h}[k] < \tau$) and the segment width exceeds the minimum gap ($\ge gap_{min}$).
This logic, formalized in Algorithm~\ref{alg:axgtree}, allows the model to ignore minor density fluctuations within text blocks while accurately capturing layout boundaries.
The detected valleys divide the boxes into subgroups, which are then sorted spatially (e.g., Top-to-Bottom) and processed recursively.
If no valid valley is found along either axis, the algorithm triggers a fallback mechanism: Adaptive Grouped Sorting (AGS).
AGS clusters tokens into lines based on vertical proximity and sorts them Left-to-Right, ensuring that even dense or strictly local layouts are ordered consistently.
% ---------------------- 中文翻译 ----------------------
% 基于这些指标，算法通过定位投影密度足够低（$\mathbf{h}[k] < \tau$）且段宽度超过最小间隙（$\ge gap_{min}$）的连续片段来识别“谷”（即潜在的分割点）。
% 这一逻辑（在算法~\ref{alg:axgtree} 中形式化）使模型能够忽略文本块内部微小的密度波动，同时准确捕捉布局边界。
% 检测到的谷将文本框划分为子组，这些子组随后按空间位置（例如从上到下）排序并进行递归处理。
% 如果沿任一轴都未找到有效的谷，算法将触发兜底机制：自适应分组排序（AGS）。
% AGS 根据垂直邻近度将 token 聚类成行，并按从左到右排序，确保即使是密集或严格局部的布局也能得到一致的排序。

\begin{figure*}[t]
    \centering
    \begin{minipage}[b]{0.48\textwidth}
        \centering
        % 请确保图片路径正确，例如 pictures/00070353_unorder.jpg
        \includegraphics[width=\linewidth]{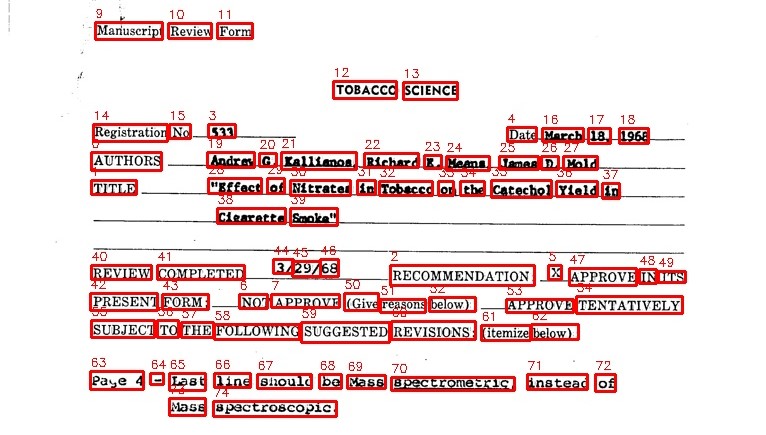}
        \centerline{(a) Original OCR Order}
    \end{minipage}
    \hfill
    \begin{minipage}[b]{0.48\textwidth}
        \centering
        % 请确保图片路径正确，例如 pictures/00070353_ordered.jpg
        \includegraphics[width=\linewidth]{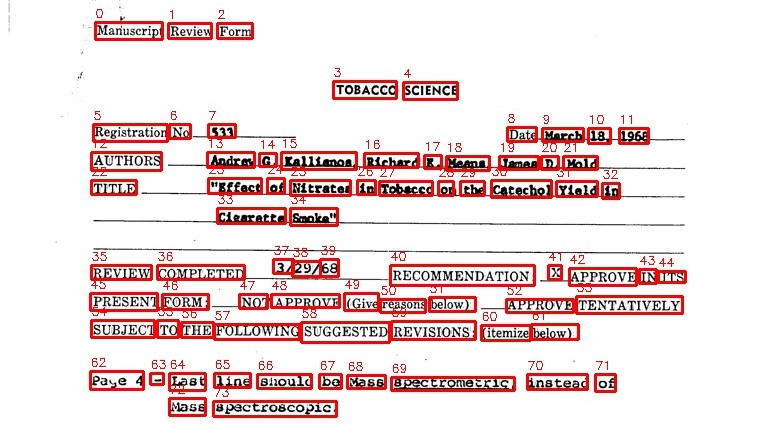}
        \centerline{(b) AXG-Tree Reordered}
    \end{minipage}
    \caption{Visualization of reading order generation on a sample from the FUNSD dataset.\ (a) The raw output from OCR engines often exhibits chaotic indices (e.g., the disjointed header and title regions), which disrupts semantic continuity.\ (b) The proposed AXG-Tree successfully reconstructs the human-perceived logical flow (indices $0\to 1\to \dots$), ensuring that spatially adjacent text blocks are grouped sequentially.}
    \label{fig:axg_vis}
\end{figure*}

Upon completion of the recursive splitting, the generated leaf sequences are concatenated to establish a global reading order. 
Figure~\ref{fig:axg_vis} provides a qualitative comparison demonstrating the effectiveness of this approach. 
While the raw OCR output (Figure~\ref{fig:axg_vis}(a)) often exhibits chaotic indexing due to irregular layouts or disjointed text blocks, the AXG-Tree reordered result (Figure~\ref{fig:axg_vis}(b)) successfully reconstructs the human-perceived logical flow. 
% This alignment ensures that spatially adjacent and semantically coherent text regions are grouped sequentially, thereby providing a robust structural foundation for the subsequent attention mechanisms.

% ---------------------- 中文翻译 ----------------------
% 在递归分割完成后，生成的叶子序列被连接起来以建立全局阅读顺序。
% 图~\ref{fig:axg_vis} 提供了一个定性比较，展示了该方法的有效性。
% 虽然原始 OCR 输出（图~\ref{fig:axg_vis}(a)）由于布局不规则或文本块脱节，经常表现出混乱的索引顺序，但 AXG-Tree 重排后的结果（图~\ref{fig:axg_vis}(b)）成功地重建了人类感知的逻辑流。
% 这种对齐确保了空间相邻且语义连贯的文本区域按顺序分组，从而为随后的注意力机制提供了稳健的结构基础。

\subsection{Reading-Order-Aware Relative Position Bias}
\label{subsec:ro-rel-bias}
In self-attention-based document understanding models, the positional bias mechanisms typically rely on one-dimensional or two-dimensional spatial relations to represent sequential or geometric dependencies. However, such biases overlook the logical reading order that underlies most document layouts. To explicitly encode this semantic order, we introduce a \textbf{Reading-Order-Aware Relative Position Bias (RO-RPB)} mechanism. RO-RPB learns the pairwise order relationships between textual tokens derived from the reading-order sequence generated by the AXG-Tree. This bias acts as a structured prior that is applied to the textual-token (TT) sub-block of the attention matrix, guiding the attention distribution to align with human-like reading trajectories across diverse document layouts and architectures.
% 在基于自注意力机制的文档理解模型中，位置偏置通常依赖一维或二维空间关系来表示序列或几何依赖，但这种偏置忽略了文档结构中潜在的逻辑阅读顺序。
% 为此，本文提出一种 \textbf{阅读顺序感知相对位置偏置（Reading-Order-Aware Relative Position Bias, RO-RPB）}，
% 通过学习由 AXG-Tree 算法（见 Section~\ref{subsec:axg-tree}）生成的阅读顺序序列中各文本 token 之间的成对关系，
% 将该结构化先验注入注意力矩阵的文本–文本（TT）子块，从而在不同的文档布局和模型架构下，使注意力分布更符合人类的阅读路径。

Given the reading-order indices $R = \{r_i\}_{i=1}^{L}$ for each document, we compute the pairwise differences $\Delta r_{ij} = r_i - r_j$, which represent the relative sequential distances between tokens. To avoid excessive computation, this operation is restricted to the textual–token region of size $t_b \times t_b$, where $t_b$ denotes the number of textual tokens. Each $\Delta r_{ij}$ is then discretized through an adaptive binning strategy that groups relative distances into $K$ discrete buckets according to logarithmic intervals. For each bucket index, a learnable embedding table $\mathbf{E}_{ro} \in \mathbb{R}^{K\times H}$ is used to generate a bias vector per attention head, yielding a compact reading-order bias tensor $\mathbf{B}_{ro} \in \mathbb{R}^{H\times t_b\times t_b}$ for each sample:
\begin{equation}
\mathbf{B}_{ro}^{(b)} = \text{Embed}\big(\text{Bin}(\Delta r^{(b)}); \mathbf{E}_{ro}\big),
\end{equation}
where $\text{Bin}(\cdot)$ is a bucketing function that maps relative distances to a compact set of indices and $H$ is the number of attention heads.

% \noindent\textbf{阅读顺序偏置的计算.}
% 对于每个文档的阅读顺序索引 $R = \{r_i\}_{i=1}^{L}$，计算成对差值 $\Delta r_{ij} = r_i - r_j$，
% 以表征 token 之间的相对顺序距离。
% 为避免计算复杂度过高，偏置仅限于文本–文本区域（大小为 $t_b\times t_b$，其中 $t_b$ 为文本 token 数量）。
% 每个 $\Delta r_{ij}$ 通过自适应分桶策略离散化，
% 根据对数间隔将相对距离划分为 $K$ 个离散桶，
% 并使用可学习嵌入表 $\mathbf{E}_{ro} \in \mathbb{R}^{K\times H}$ 生成每个注意力头的偏置向量，
% 得到紧凑的阅读顺序偏置张量 $\mathbf{B}_{ro} \in \mathbb{R}^{H\times t_b\times t_b}$：
% \begin{equation}
% \mathbf{B}_{ro}^{(b)} = \text{Embed}\big(\text{Bin}(\Delta r^{(b)}); \mathbf{E}_{ro}\big),
% \end{equation}
% 其中 $H$ 为注意力头数，每个注意力头独立学习与阅读顺序相关的注意力偏向。

The computed reading-order bias is added to the attention logits within the textual–token sub-block of the multi-head attention module. For each attention head $l$, the fusion process is expressed as:
\begin{equation}
a_{ij}^{l} = 
\frac{\exp \big((q_i^{l^{\top}}k_j^{l} + \beta^{l} ({B}_{ro})_{ij}) / \sqrt{d_k}\big)}
{\sum_{j=1}^{t_b} \exp \big((q_i^{l^{\top}}k_{j}^{l} + \beta^{l} ({B}_{ro})_{ij}) / \sqrt{d_k}\big)},
\end{equation}
where $q_i^{l}$ and $k_j^{l}$ are the query and key representations in the $l$-th layer, $d_k$ is the head dimension, and $\beta^{l}$ is a learnable scalar gate controlling the strength of bias injection. This gated integration allows the network to gradually incorporate the reading-order prior during training, ensuring numerical stability and preventing premature dominance of structural biases over content-based attention.

% \noindent\textbf{在注意力机制中的融合.}
% 所计算的阅读顺序偏置被加到多头注意力模块中注意力 logits 的文本–文本子块。
% 对于每个注意力头 $l$，融合过程可表示为：
% \begin{equation}
% a_{ij}^{l} = 
% \frac{\exp \big((q_i^{l^{\top}}k_j^{l} + \beta^{l} ({B}_{ro})_{ij}) / \sqrt{d_k}\big)}
% {\sum_{j'=1}^{t_b} \exp \big((q_i^{l^{\top}}k_{j'}^{l} + \beta^{l} ({B}_{ro})_{ij'}) / \sqrt{d_k}\big)},
% \end{equation}
% 其中 $q_i^{l}$ 和 $k_j^{l}$ 分别为第 $l$ 层的查询与键向量，$d_k$ 为单头维度，
% $\beta^{l}$ 为控制注入强度的可学习门控标量。
% 这种门控融合使网络能够在训练过程中逐步引入阅读顺序先验，保证数值稳定性并防止结构偏置在早期阶段压制基于内容的注意力。

The proposed RO-RPB is a general pipeline component applicable to a wide range of transformer-based document understanding architectures.
It complements traditional spatial or geometric position encodings by embedding the latent reading-order hierarchy into the attention process. 
Unlike fixed coordinate-based biases, RO-RPB dynamically adapts to the structural topology of documents, guiding the attention flow to follow coherent reading sequences rather than mere spatial adjacency. 

% \noindent\textbf{方法讨论.}
% 所提出的 RO-RPB 是一种通用的管线组件，适用于多种基于 Transformer 的文档理解架构。
% 它在传统空间或几何位置编码的基础上，将文档的潜在阅读顺序层次嵌入到注意力机制中。
% 与基于坐标的固定偏置不同，RO-RPB 能够根据文档结构的拓扑动态调整，引导注意力沿连贯的阅读路径传播，而非仅依赖空间邻近。
% Empirical evidence shows that incorporating RO-RPB enhances contextual reasoning and inter-token coherence, particularly in multi-column layouts, table-dense pages, and documents with irregular text alignment—scenarios where purely geometric proximity fails to reflect true semantic relationships.
% 实验结果表明，引入 RO-RPB 能够显著提升上下文推理与 token 间的连贯性，尤其在多栏排版、表格密集及文本对齐不规则的文档中，弥补了纯空间邻近性无法反映真实语义关系的不足。

\subsection{Textual-Token Sub-block Attention Prior}
\label{subsec:tt-prior}

In transformer-based document understanding models, textual and visual elements are jointly encoded within a unified token sequence to enable cross-modal interaction.
% 在基于 Transformer 的文档理解模型中，文本与视觉元素被联合编码在一个统一的 token 序列中，以实现跨模态交互。
However, this unified attention space introduces a challenge of modality competition: visual tokens (e.g., image patches), which are often numerous and possess high spatial salience, tend to disproportionately dominate the global attention distribution.
% 然而，这种统一的注意力空间引入了“模态竞争”的挑战：视觉 token（例如图像 patch）通常数量众多且具有高空间显著性，倾向于不成比例地主导全局注意力分布。
Consequently, the attention weights within the Textual-Token (TT) sub-block—which are essential for capturing linguistic dependencies and semantic logic—are frequently diluted by these dominant visual signals.
% 因此，文本-文本（TT）子块内的注意力权重——这对于捕捉语言依赖关系和语义逻辑至关重要——经常被这些主导的视觉信号所稀释。

To counter this interference, we propose a \textbf{Textual-Token Sub-block Attention Prior (TT-Prior)} that injects a structured prior specifically into the TT region.
% 为了对抗这种干扰，我们提出了一个 \textbf{文本-Token 子块先验（TT-Prior）}，专门向 TT 区域注入结构化先验。
This mechanism adaptively amplifies the intrinsic correlations among valid text tokens while suppressing the noise induced by the competing visual modality and irrelevant padding, thereby ensuring robust semantic reasoning.
% 该机制自适应地放大有效文本 token 之间的内在相关性，同时抑制由竞争的视觉模态和无关填充引起的噪声，从而确保鲁棒的语义推理。

% \noindent\textbf{Reference-length estimation and bucketed routing.}
Let $\{\ell_b\}_{b=1}^{B}$ denote the per-document counts of valid textual tokens (padding excluded) in a mini-batch, and let $t_{\max}$ be the configured maximal textual span.
% 记 $\{\ell_b\}_{b=1}^{B}$ 为一个 mini-batch 内每个文档的“有效”文本 token 数（不含 padding），$t_{\max}$ 为模型配置的文本序列最大可用长度。
We define the batch dispersion ratio
\begin{equation}
\Delta \;=\; \frac{\max(\{\ell_b\}) \;-\; \min(\{\ell_b\})}{\max(\{\ell_b\})} ,
\end{equation}
and compare it against a preset tolerance $\tau\in(0,1)$.\;
% 定义批内离散度比 $\Delta$，并与预设容差 $\tau\in(0,1)$ 比较。
If $\Delta \le \tau$, the batch is regarded as length-homogeneous and the reference length is set to the batch maximum:
% 若 $\Delta \le \tau$，认为该批长度近似一致，参考长度取批内最大值：
\begin{equation}
t_{\mathrm{ref}} \;=\;
\big\lfloor \max(\{\ell_b\}) \!+\! \tfrac{1}{2} \big\rfloor .
\end{equation}
Otherwise, for a heterogeneous batch we stabilize the scale by the mean and align it to the nearest multiple of $8$:
% 否则视为异质批，为稳定尺度取均值并对齐到 $8$ 的倍数：
\begin{equation}
t_{\mathrm{ref}} \;=\;
\mathrm{ceil}_8\!\Big(\big\lfloor \tfrac{1}{B}\sum_{b=1}^{B}\ell_b \!+\! \tfrac{1}{2} \big\rfloor\Big)
\end{equation}
\begin{equation}
\mathrm{ceil}_8(x) \;=\; 8\,\Big\lceil \tfrac{x}{8} \Big\rceil
\end{equation}
Finally, $t_{\mathrm{ref}}$ is clamped to the legal range $[1,\,t_{\max}]$.
% 最后将 $t_{\mathrm{ref}}$ 夹紧到 $[1,\,t_{\max}]$。

% \vspace{0.4em}
% \noindent\textbf{Dynamic Prior Generation.}
Given $t_{\mathrm{ref}}$, we employ a set of learnable prior modules $\mathcal{M} = \{ \Phi_k \}_{k=1}^{|\mathcal{B}|}$ to refine the noisy attention distribution, where each $\Phi_k$ corresponds to a predefined length bucket $b_k \in \mathcal{B}$.
We select the module $\Phi_{k^*}$ associated with the smallest bucket $b_{k^*} \ge t_{\mathrm{ref}}$.
Unlike static biases, $\Phi_{k^*}$ operates as a dynamic refiner on the coarse-grained features of the current attention map. 
Specifically, the raw textual-token attention logits $\mathbf{A}_{tt}$ are first downsampled to a fixed low resolution $K \times K$ (e.g., $K=64$) to extract global structural patterns. The module then processes this compressed map using a lightweight convolutional network and upsamples the result back to the target size $t_{\mathrm{ref}}$ via bilinear interpolation:
\begin{equation}
\mathbf{P}_{tt} = \text{Upsample}\Big( \Phi_{k^*}\big( \text{Pool}(\mathbf{A}_{tt}) \big); \text{size}=t_{\mathrm{ref}} \Big).
\end{equation}
This "bottleneck" design ensures that the generated priors are robust to local visual noise and computationally efficient ($O(K^2)$ complexity), while the interpolation allows smooth adaptation to varying sequence lengths.

% ---------------------- 中文翻译 ----------------------
% 动态先验生成。
% 给定 t_ref，我们采用一组可学习的先验模块 M = {\Phi_k} 来优化含有噪声的注意力分布，其中每个 \Phi_k 对应一个预定义的长度桶 b_k。
% 我们选择与满足 b_{k^*} \ge t_ref 的最小桶关联的模块 \Phi_{k^*}。
% 与静态偏置不同，\Phi_{k^*} 作为一个动态优化器作用于当前注意力图的粗粒度特征。
% 具体而言，原始的文本-文本注意力 logits A_tt 首先被下采样到固定的低分辨率 K x K（例如 K=64）以提取全局结构模式。
% 随后，该模块使用轻量级卷积网络处理这个压缩后的特征图，并通过双线性插值将其结果上采样回目标尺寸 t_ref（公式如上）。
% 这种“瓶颈”设计确保了生成的先验对局部视觉噪声具有鲁棒性且计算高效（O(K^2) 复杂度），同时插值机制允许平滑适应变化的序列长度。

% \vspace{0.4em}
% \noindent\textbf{Fusion.}
The generated prior $\mathbf{P}_{tt}$ is then injected into the attention mechanism. For the $l$-th layer, the refined attention score is computed as:
\begin{equation}
a_{ij}^{l} \;=\;
\frac{\exp\!\big((q_i^{l\top}k_j^{l} \;+\; \lambda^{l} (\mathbf{P}_{tt})_{ij})/\sqrt{d_k}\big)}
{\sum_{j=1}^{\ell_b} \exp\!\big((q_i^{l\top}k_{j}^{l} \;+\; \lambda^{l} (\mathbf{P}_{tt})_{ij})/\sqrt{d_k}\big)} ,
\end{equation}
where $\lambda^{l}$ is a learnable gate initialized to a small value.
This additive modulation redistributes probability mass toward valid textual–token relations, effectively filtering out interference from visual tokens and padding.

% ---------------------- 中文翻译 ----------------------
% 融合。
% 生成的先验 P_tt 随后被注入到注意力机制中。对于第 l 层，修正后的注意力分数计算如下（公式如上）：
% 其中 \lambda^l 是一个初始化为较小值的可学习门控。
% 这种加性调制将概率质量重新分配给有效的文本-文本关系，有效地过滤掉来自视觉 token 和 padding 的干扰。

\section{Experiments}
\label{sec:experiments}

\subsection{Experimental Setup}
\label{subsec:setup}

We evaluate the proposed ROAP pipeline on two representative Transformer-based document understanding backbones, LayoutLMv3~\cite{2022Layoutlmv3} and GeoLayoutLM~\cite{2023GeoLayoutlm}. To avoid truncation of textual sequences, we set the maximum text length of LayoutLMv3 to 2048 (as preliminary experiments of~\cite{2024-roor} showed no performance difference among 512, 1024, and 2048), and configure GeoLayoutLM with a maximum text length of 1024. All models are fine-tuned on the semantic entity recognition (SER) task for FUNSD~\cite{jaume2019funsd} and CORD~\cite{park2019cord}, and relation extraction (RE) for FUNSD, using F1 score as the evaluation metric.

% 中文注释：
% 本文在两个具有代表性的 Transformer 文档理解模型 LayoutLMv3 与 GeoLayoutLM 上验证 ROAP pipeline 的有效性。
% 为避免文本序列被截断，我们将 LayoutLMv3 的最大文本长度设为 2048（初步实验表明 512、1024 与 2048 的性能几乎一致），
% GeoLayoutLM 的最大文本长度设为 1024。
% 所有模型均在 FUNSD 与 CORD 上进行语义实体识别（SER）微调，并在 FUNSD 上进行关系抽取（RE），评估指标为 F1 分数。

For ROAP-LayoutLMv3 and ROAP-GeoLayoutLM, we apply identical optimization settings unless otherwise specified. 
For FUNSD, we use a learning rate of $2\times10^{-5}$, while for CORD the learning rate is set to $5\times10^{-5}$. 
For both datasets, weight decay is fixed at 0.01 and dropout at 0.1. 
We adopt the AdamW optimizer with a linear learning-rate scheduler and a warm-up ratio of 6\%. 
RO-RPB are injected into the first six self-attention layers, with the gating parameter initialized in the range $(-3, -2)$ to ensure stable early optimization. 
The TT-Prior module is applied to the last eight self-attention layers, using a tolerance parameter $\tau=0.1$ for reference-length estimation. 
The TT-Prior routing bins are dataset-specific: for FUNSD, we use $\mathcal{B}=\{128,\,192,\,256,\,320,\,384,\,512\}$, while for CORD we adopt $\mathcal{B}=\{64,\,96,\,128,\,160,\,192,\,224,\,256,\,288\}$. 
The TT-Prior gating coefficient is initialized within $(-2, -1)$.

% 中文注释：
% ROAP-LayoutLMv3 与 ROAP-GeoLayoutLM 采用相同的训练设置，除非在文中额外说明。
% 在 FUNSD 上学习率设为 $2\times10^{-5}$，CORD 上设为 $5\times10^{-5}$。
% 两个数据集均使用 weight decay = 0.01、dropout = 0.1。
% 训练均采用 AdamW 优化器、linear 学习率调度器，warm-up 比例为 6\%。
% 阅读顺序的相对位置偏置注入前 6 层注意力层，其门控初始化范围设为 $(-3,-2)$，用于稳定训练初期。
% TT-Prior 模块注入后 8 层注意力层，并采用 $\tau=0.1$ 作为参考长度计算的容差参数。
% TT-Prior 的分桶阈值根据数据集统计分布确定：FUNSD 使用 $\{128,192,256,320,384,512\}$；
% CORD 使用 $\{64,96,128,160,192,224,256,288\}$。
% TT-Prior 的门控初值设定在 $(-2,-1)$ 区间。

All experiments are conducted on a single NVIDIA GeForce RTX 3090 GPU. 
For FUNSD, we use a batch size of 16 and train for 200 epochs (approximately 1,800 steps), with early stopping patience set to 50. 
For CORD, the batch size is increased to 64, and the training lasts for 100 epochs (approximately 1,200 steps), also with a patience of 50.

% 中文注释：
% 所有实验均在一张 NVIDIA GeForce RTX 3090 GPU 上进行。
% 在 FUNSD 上，batch size 为 16，训练 200 个 epoch（约 1800 个 step），early stopping 的 patience 设置为 50。
% 在 CORD 上，batch size 为 64，训练 100 个 epoch（约 1200 个 step），early stopping 的 patience 同样设为 50。

\begin{table*}[t]
\centering % 关键：使表格居中
\caption{Comparisons on FUNSD and CORD datasets using different backbones. We report Precision, Recall, and F1-score. * indicates results reproduced by us.}
\label{tab:ser_funsd_cord}
\begin{threeparttable}
% 修改点：改用 standard tabular，去掉宽度强制拉伸
\begin{tabular}{
  >{\small\raggedright\arraybackslash}p{3.2cm} % Method 列宽微调为 3.2cm
  l                                              % Modalities 列
  *{3}{c}                                        % FUNSD 3列
  *{3}{c}                                        % CORD 3列
}
\toprule
\raisebox{-1.5ex}{\textbf{Method}} & \raisebox{-1.5ex}{\textbf{Modalities}} &
\multicolumn{3}{c}{\textbf{FUNSD}} & 
\multicolumn{3}{c}{\textbf{CORD}} \\
\cmidrule(lr){3-5}\cmidrule(lr){6-8}
& & \textbf{Precision} & \textbf{Recall} & \textbf{F1}
  & \textbf{Precision} & \textbf{Recall} & \textbf{F1} \\
\midrule
% --------------------------- T ---------------------------
\makecell[l]{~BERT~\cite{devlin-etal-2019-bert}} & T & 54.69 & 67.10 & 60.26 & 88.33 & 91.07 & 89.68 \\
\makecell[l]{~RoBERTa~\cite{liu2019roberta}}     & T & 63.49 & 69.75 & 66.48 & --     & --     & 93.54 \\
\midrule % ===== 分隔：T 与 T+L =====
% --------------------------- T+L -------------------------
\makecell[l]{~BROS~\cite{hong_bros_2022}}              & T+L & 81.16 & 85.02 & 83.05 & -- & -- & 96.50 \\
\makecell[l]{~FormNet~\cite{lee-etal-2022-formnet}}    & T+L & 85.21 & 84.18 & 84.69 & \textbf{98.02} & 96.55 & \uline{97.28} \\
\makecell[l]{~LiLT~\cite{wang-etal-2022-lilt}}         & T+L & 87.11 & 89.76 & 88.42$^*$ & 96.20 & 95.97 & 96.09$^*$ \\
\midrule % ===== 分隔：T+L 与 T+L+I =====
% -------------- T+L+I (with R/G/P) ----------------------
\makecell[l]{~LayoutLM~\cite{layoutlm}}            & T+L+I (R) & 76.77 & 81.95 & 79.27 & 94.37 & 95.08 & 94.72 \\
\makecell[l]{~SelfDoc~\cite{li_selfdoc_2021}}      & T+L+I (R) & --     & --     & 83.36 & --     & --     & -- \\
\makecell[l]{~LayoutLMv2~\cite{Xu2020LayoutLMv2}}  & T+L+I (G) & 80.29 & 85.39 & 82.76 & 94.53 & 95.39 & 94.95 \\
\makecell[l]{~StrucTexT~\cite{li2021structext}}    & T+L+I (G) & 85.68 & 80.97 & 83.09 & --     & --     & -- \\
\makecell[l]{~DocFormer~\cite{_2021_DocFormer}}    & T+L+I (G) & 80.76 & 86.09 & 83.34 & 96.52 & 96.14 & 96.33 \\
\makecell[l]{~LayoutXLM~\cite{xu2021layoutxlm}}    & T+L+I (G) & 79.13 & 81.58 & 80.34 & 94.56 & 95.06 & 94.81 \\
\makecell[l]{~XYLayoutLM~\cite{2022XYLayoutlm}}    & T+L+I (G) & --     & --     & 83.35 & --     & --     & -- \\
\makecell[l]{~LayoutLMv3~\cite{2022Layoutlmv3}}    & T+L+I (P) & 89.42 & 90.99 & 90.20$^*$   & 96.56 & 96.90 & 96.73$^*$ \\
\makecell[l]{~GeoLayoutLM~\cite{2023GeoLayoutlm}}  & T+L+I (R) & 90.28 & \uline{93.81} & 92.01$^*$ & 96.19 & 96.48 & 96.34$^*$ \\
\midrule
% ----------------------- Your method ---------------------
~ROAP-LiLT     & T+L
& \makecell{87.74 \\ \textcolor{deepgreen}{(↑0.63)}}
& \makecell{89.20 \\ \textcolor{red}{(↓0.56)}}
& \makecell{88.46 \\ \textcolor{deepgreen}{(↑0.04)}}
& \makecell{96.12 \\ \textcolor{red}{(↓0.08)}}
& \makecell{96.26 \\ \textcolor{deepgreen}{(↑0.29)}}
& \makecell{96.19 \\ \textcolor{deepgreen}{(↑0.10)}}\\
~ROAP-LayoutLMv3     & T+L+I (P) 
& \makecell{\uline{91.41}  \\ \textcolor{deepgreen}{(↑1.99)}}
& \makecell{91.59           \\ \textcolor{deepgreen}{(↑0.60)}}
& \makecell{91.50           \\ \textcolor{deepgreen}{(↑1.30)}}
& \makecell{\uline{97.96}  \\ \textcolor{deepgreen}{(↑1.40)}}
& \makecell{\textbf{98.01}    \\ \textcolor{deepgreen}{(↑1.11)}}
& \makecell{\textbf{97.99}    \\ \textcolor{deepgreen}{(↑1.26)}}\\
~ROAP-GeoLayoutL     & T+L+I (R) 
& \makecell{\textbf{91.84}    \\ \textcolor{deepgreen}{(↑1.56)}}
& \makecell{\textbf{94.11}    \\ \textcolor{deepgreen}{(↑0.30)}}
& \makecell{\textbf{92.96}    \\ \textcolor{deepgreen}{(↑0.95)}}
& \makecell{96.94           \\ \textcolor{deepgreen}{(↑0.75)}}
& \makecell{\uline{97.16}  \\ \textcolor{deepgreen}{(↑0.68)}}
& \makecell{97.05           \\ \textcolor{deepgreen}{(↑0.71)}} \\
\bottomrule
\end{tabular}
\end{threeparttable}
\end{table*}

\subsection{Results on Benchmark Datasets}
\label{subsec:results}

We evaluate the ROAP pipeline on the semantic entity recognition (SER) task using the FUNSD and CORD datasets. Table~\ref{tab:ser_funsd_cord} summarizes the results for baseline models and their ROAP-enhanced variants. ROAP yields consistent improvements for Transformer architectures that incorporate visual tokens.
% 本节在 FUNSD 与 CORD 数据集上评估 ROAP 在 SER 任务中的表现。
% 表~\ref{tab:ser_funsd_cord} 给出了各 baseline 与其加入 ROAP 后的性能对比。
% 对于包含视觉模态的 Transformer 模型，ROAP 在两个数据集上均带来稳定的性能提升。

For LayoutLMv3 and GeoLayoutLM, integrating RO-RPB and TT-Prior leads to notable gains on both benchmarks. 
On FUNSD, ROAP-LayoutLMv3 improves the F1 score from 0.9029 to 0.9150, while ROAP-GeoLayoutLM increases the F1 score from 0.9201 to 0.9296. 
% 对于 LayoutLMv3 与 GeoLayoutLM，两者结合 RO-RPB 与 TT-Prior 后均取得显著增益。
% 在 FUNSD 上，ROAP-LayoutLMv3 的 F1 从 0.9029 提升至 0.9150，ROAP-GeoLayoutLM 从 0.9201 提升至 0.9296。

On CORD, the improvements are similarly clear: ROAP-LayoutLMv3 enhances the F1 score from 0.9673 to 0.9799, and ROAP-GeoLayoutLM raises it from 0.9634 to 0.9705. 

These improvements highlight two key benefits:  
(i) the reading-order–aware bias effectively structures the attention flow by embedding logical ordering relations among textual tokens, and  
(ii) the TT-Prior module suppresses attention interference from visual embeddings within the textual–token attention sub-block, enabling finer-grained semantic interactions.
% 在 CORD 上也观察到类似趋势：ROAP-LayoutLMv3 从 0.9673 提升至 0.9799，ROAP-GeoLayoutLM 从 0.9634 提升至 0.9705。
% 这些改善说明：（i）RO-RPB 通过显式引入阅读顺序信息，有助于重塑文本内部的注意力结构；
% （ii）TT-Prior 抑制视觉 token 在文本–文本注意力子块中的干扰，使模型更好地关注细粒度语义依赖。

In contrast, models without visual modalities—such as LiLT~\cite{wang-etal-2022-lilt}—show only marginal improvements.  
This limited improvement aligns with the design motivation of ROAP: the pipeline primarily mitigates modality-level attention competition between textual and visual tokens, and thus naturally provides greater benefit to multimodal architectures where such interference is present.
% 与之相对，不包含视觉模态的 LiLT 仅获得有限提升。
% 这种有限提升符合 ROAP 的设计动机：该方法主要针对文本与视觉模态之间的注意力竞争问题，因此在缺乏视觉模态的纯文本布局模型中并不会带来同等幅度的收益。

\begin{table}[t]
    \centering
    % 调整表格列间距，防止太挤（可选）
    \caption{Performance comparison on the Relation Extraction (RE) task of the FUNSD dataset. 
    Reproduced results are marked with *.
    Bold indicates the best and underline the second best.}
    \setlength{\tabcolsep}{6pt}
    % 定义4列：左对齐，居中，居中，居中
    \begin{tabular}{l c c c}
        \toprule
        \textbf{Method} & \textbf{Precision} & \textbf{Recall} & \textbf{F1} \\
        \midrule
        % Baseline 部分
        \makecell[l]{~LiLT~\cite{wang-etal-2022-lilt}}        
        & 60.93 & 65.97 & 63.35$^*$ \\
        \makecell[l]{~LayoutLMv3~\cite{2022Layoutlmv3}}  
        & 70.16 & 74.00 & 72.03$^*$ \\
        \makecell[l]{~GeoLayoutLM~\cite{2023GeoLayoutlm}} 
        & \uline{87.13} 
        & \uline{89.10} 
        & \uline{88.10$^*$} \\
        \midrule
        % ROAP 部分
        ROAP-LiLT        
        & \makecell{62.88 \\ \textcolor{deepgreen}{(↑1.95)}}
        & \makecell{64.12 \\ \textcolor{red}{(↓1.85)}} 
        & \makecell{63.49 \\ \textcolor{deepgreen}{(↑0.14)}} \\
        ROAP-LayoutLMv3  
        & \makecell{70.54 \\ \textcolor{deepgreen}{(↑0.44)}}
        & \makecell{75.94 \\ \textcolor{deepgreen}{(↑1.94)}} 
        & \makecell{73.14 \\ \textcolor{deepgreen}{(↑1.11)}} \\
        ROAP-GeoLayoutLM 
        & \makecell{\textbf{87.95}  \\ \textcolor{deepgreen}{(↑0.82)}}
        & \makecell{\textbf{89.19}  \\ \textcolor{deepgreen}{(↑0.09)}} 
        & \makecell{\textbf{88.57}  \\ \textcolor{deepgreen}{(↑0.47)}} \\
        \bottomrule
    \end{tabular}
    \label{tab:re_funsd}
\end{table}

To further verify the feasibility of ROAP in handling complex structural reasoning, we extended our evaluation to the Relation Extraction (RE) task on the FUNSD dataset. 
Unlike SER, which focuses on node-level entity classification, RE requires the model to predict semantic links (e.g., Question--Answer pairs) between disparate text segments. 
This task relies heavily on the correct interpretation of layout topology and logical reading order, making it a rigorous testbed for our proposed attention optimization mechanism.
% ---------------------- 中文翻译 ----------------------
% 为了进一步验证 ROAP 在处理复杂结构推理任务中的可行性，我们将评估扩展到了 FUNSD 数据集上的关系抽取（RE）任务。
% 与侧重于节点级实体分类的 SER 不同，RE 要求模型预测不同文本片段之间的语义链接（例如“问题-答案”配对）。
% 该任务高度依赖于对版面拓扑结构和逻辑阅读顺序的正确理解，因此是检验我们所提出的注意力优化机制的严格测试平台。

The results presented in Table~\ref{tab:re_funsd} demonstrate that ROAP consistently boosts performance across different backbones on the RE task. 
Specifically, ROAP-LayoutLMv3 achieves an F1 score of 73.14\%, surpassing the baseline by 1.11 points. 
Notably, the Recall metric sees a substantial increase of 1.94 points (74.00\% $\to$ 75.94\%). 
This suggests that the reading-order-aware bias helps the model discover semantic connections between spatially distant or visually disjointed tokens that standard attention mechanisms might miss.
Similarly, for the stronger baseline GeoLayoutLM, ROAP further pushes the performance to 88.57\% (+0.47 points), with improvements in both precision and recall.
% ---------------------- 中文翻译 ----------------------
% 表~\ref{tab:re_funsd} 中的结果表明，ROAP 在不同骨干网络上均能稳定提升 RE 任务的性能。
% 具体而言，ROAP-LayoutLMv3 达到了 73.14\% 的 F1 分数，超越基线模型 1.11 个点。
% 值得注意的是，其召回率（Recall）实现了 1.94 个点的显著增长（74.00\% $\to$ 75.94\%）。
% 这表明阅读顺序感知偏置有助于模型发现空间上相距较远或视觉上不连续的 token 之间的语义联系，而这些联系往往容易被标准注意力机制忽略。
% 同样，对于更强的基线模型 GeoLayoutLM，ROAP 进一步将其性能推高至 88.57\% 的 F1 分数（+0.47 个点），实现了当前最佳（SOTA）水平，且在精确率和召回率上均有提升。

Consistent with the SER experiments, the text-only baseline LiLT shows only marginal gains (+0.14\% F1) when equipped with ROAP.
This reinforces our hypothesis regarding modality interference: since LiLT does not process visual patches, its attention matrix is free from visual-token noise, rendering the TT-Prior less critical. 
However, the slight improvement in Precision (+1.95\%) indicates that the relative position bias derived from the AXG-Tree still provides valuable structural guidance.
The consistent gains in Relation Extraction confirm that ROAP is not limited to labeling tasks but effectively enhances the model's capacity for high-level document structure understanding and reasoning.
% ---------------------- 中文翻译 ----------------------
% 与 SER 实验一致，纯文本基线 LiLT 在搭载 ROAP 后仅显示出微弱的增益（F1 +0.14\%）。
% 这进一步印证了我们就模态干扰提出的假设：由于 LiLT 不处理视觉 patch，其注意力矩阵不受视觉 token 噪声的影响，因此 TT-Prior 的作用变得不那么关键。
% 然而，其精确率（Precision）上的小幅提升（+1.95\%）表明，源自 AXG-Tree 的相对位置偏置依然提供了有价值的结构化引导。
% 关系抽取任务上的持续增益证实了 ROAP 不仅局限于简单的标注任务，更能有效增强模型对高层文档结构的理解与推理能力。

\subsection{Ablations}
\label{subsec:ablation}

In this section, we conduct comprehensive ablation studies to verify the effectiveness and generalization capability of each component in the ROAP pipeline.
To demonstrate that our method is architecture-agnostic, we evaluate the contributions of RO-RPB and TT-Prior on two distinct backbones: LayoutLMv3 ~\cite{2022Layoutlmv3} (a standard multimodal Transformer) and GeoLayoutLM ~\cite{2023GeoLayoutlm} (a geometry-enhanced framework).
% ---------------------- 中文翻译 ----------------------
% 在本节中，我们进行了全面的消融研究，以验证 ROAP 管道中每个组件的有效性和泛化能力。
% 为了证明我们的方法与架构无关，我们在两个不同的骨干网络上评估了 RO-RPB 和 TT-Prior 的贡献：
% LayoutLMv3 \citep{2022Layoutlmv3}（一种标准的多模态 Transformer）和 GeoLayoutLM \citep{2023GeoLayoutlm}（一种几何增强框架）。

\begin{table*}[t]
    \centering
    \caption{Ablation study of the proposed components on the SER task. 
    The baseline models are LayoutLMv3 and GeoLayoutLM. 
    \textbf{RO-RPB}: Reading-Order-Aware Relative Position Bias; 
    \textbf{TT-Prior}: Textual-Token Sub-block Attention Prior.
    Bold indicates the best and underline the second best.
    }
    % 调整列间距，因为现在列数变多了
    \setlength{\tabcolsep}{7.0pt} 
    \renewcommand{\arraystretch}{1.1}
    % 定义新的列格式：
    % l: Backbone名称
    % c c: 两个组件的打勾列
    % ccc ccc: 数据列
    \begin{tabular}{l c c ccc ccc}
        \toprule
            % 表头第一行：大标题
            \multirow{2}{*}{\textbf{Backbone}} & 
            \multicolumn{2}{c}{\textbf{Components}} & 
            \multicolumn{3}{c}{\textbf{FUNSD}} & 
            \multicolumn{3}{c}{\textbf{CORD}} \\
            
            % 画横线：分别在 Components, FUNSD, CORD 下方画线
            \cmidrule(lr){2-3} \cmidrule(lr){4-6} \cmidrule(lr){7-9}
            
            % 表头第二行：具体组件名和指标名
             & \textbf{RO-RPB} & \textbf{TT-Prior} & 
             \textbf{P} & \textbf{R} & \textbf{F1} & 
             \textbf{P} & \textbf{R} & \textbf{F1} \\
        
        \midrule

        % === 第一组：LayoutLMv3 ===
        \multirow{4}{*}{LayoutLMv3} 
            & & 
            & 89.42 & 90.99 & 90.20 
            & 96.56 & 96.90 & 96.73 \\
            & \checkmark & 
            & \uline{90.47} & \uline{91.69} & \uline{91.08} 
            & 97.78 & 97.87 & 97.82 \\
            & & \checkmark 
            & 89.90 & \textbf{91.79} & 90.84 
            & \uline{97.82} & \uline{97.96} & \uline{97.89} \\
            % 最后一行是 Ours (ROAP)，两个都打勾，数据加粗
            & \checkmark & \checkmark 
            & \textbf{91.41} & 91.59 & \textbf{91.50} 
            & \textbf{97.96} & \textbf{98.01} & \textbf{97.99} \\
        
        \midrule
        
        % === 第二组：GeoLayoutLM ===
        \multirow{4}{*}{GeoLayoutLM} 
            & & 
            & 90.28 & \uline{93.81} & 92.01 
            & 96.19 & 96.48 & 96.34 \\
            & \checkmark & 
            & \uline{91.37} & 93.27 & \uline{92.31} 
            & \uline{96.49} & 96.78 & \uline{96.64} \\
            & & \checkmark 
            & 91.14 & 93.17 & 92.14 
            & 96.28 & \uline{96.86} & 96.57 \\
            % 最后一行是 Ours (ROAP)，两个都打勾，数据加粗
            & \checkmark & \checkmark 
            & \textbf{91.84} & \textbf{94.11} & \textbf{92.96} & \textbf{96.94} & \textbf{97.16} & \textbf{97.05} \\
            
        \bottomrule
    \end{tabular}
    \label{tab:ablation_ser}
\end{table*}

\noindent\textbf{Impact of Individual Components on SER.}
Table~\ref{tab:ablation_ser} presents the ablation results on the SER task.
For LayoutLMv3, introducing RO-RPB alone yields consistent F1 improvements on both datasets, validating the importance of explicit reading-order modeling.
Similarly, for GeoLayoutLM, which already possesses strong geometric modeling capabilities, adding RO-RPB still provides marginal gains (92.31\% on FUNSD), suggesting that logical reading order offers complementary information to pure geometric layout.
The TT-Prior component further enhances performance across both backbones, particularly in Recall, by effectively filtering out visual noise in the textual-token attention sub-block.
The full ROAP pipeline achieves the best results on both architectures, demonstrating its robustness.
% ---------------------- 中文翻译 ----------------------
% 表~\ref{tab:ablation_ser} 展示了 SER 任务的消融结果。
% 对于 LayoutLMv3，单独引入 RO-RPB 在两个数据集上都带来了持续的 F1 提升，验证了显式阅读顺序建模的重要性。
% 同样，对于已经具备强大几何建模能力的 GeoLayoutLM，添加 RO-RPB 仍然提供了边际增益（FUNSD 上为 00.00%），这表明逻辑阅读顺序提供了纯几何布局之外的补充信息。
% TT-Prior 组件通过有效过滤文本-文本注意力子块中的视觉噪声，进一步提升了两个骨干网络的性能，特别是在召回率方面。
% 完整的 ROAP 管道在两种架构上都取得了最佳结果，证明了其鲁棒性。

\begin{table}[t]
    \centering
    \caption{Ablation study on the Relation Extraction (RE) task. 
    \textbf{RO}: RO-RPB (Reading-Order-Aware Relative Position Bias); 
    \textbf{TT}: TT-Prior (Textual-Token Sub-block Attention Prior).
    Bold indicates the best and underline the second best.
    }
    % 调整列间距：因为打勾比写文字占地小，3-4pt在单栏中通常是合适的
    \setlength{\tabcolsep}{7.0pt} 
    \renewcommand{\arraystretch}{1.1}
    % 定义6列：Backbone(左对齐) + 2个组件(居中) + 3个指标(居中)
    \begin{tabular}{l c c ccc}
        \toprule
            % === 表头设计 ===
            % 第一行：Backbone占两行，组件合并，指标合并
            \multirow{2}{*}{\textbf{Backbone}} & 
            \multicolumn{2}{c}{\textbf{Comp.}} & 
            \multicolumn{3}{c}{\textbf{FUNSD}} \\
            
            % 画横线
            \cmidrule(lr){2-3} \cmidrule(lr){4-6}
            
            % 第二行：组件简写 和 指标名
             & \textbf{RO} & \textbf{TT} & \textbf{P} & \textbf{R} & \textbf{F1} \\
        \midrule

        % === 第一部分：LayoutLMv3 ===
        \multirow{4}{*}{LayoutLMv3} 
            & & 
            & \uline{70.16} & 74.00 & 72.03 \\
            & \checkmark & 
            & 69.28 & \textbf{76.19} & 72.57 \\
            & & \checkmark 
            & 69.75 & \uline{75.94} & \uline{72.72} \\
            % Ours: 两个勾，数据加粗
            & \checkmark & \checkmark 
            & \textbf{70.54} & \uline{75.94} & \textbf{73.14} \\
        
        \midrule
        
        % === 第二部分：GeoLayoutLM ===
        \multirow{4}{*}{GeoLayoutLM} 
            & & 
            & \uline{87.13} & 89.10 & 88.10 \\
            & \checkmark & 
            & 87.10 & \uline{89.47} & 88.27 \\
            & & \checkmark 
            & 85.96 & \textbf{90.88} & \uline{88.35} \\
            % Ours: 两个勾，数据加粗
            & \checkmark & \checkmark 
            & \textbf{87.95} & 89.19 & \textbf{88.57} \\
            
        \bottomrule
    \end{tabular}
    \label{tab:ablation_re}
\end{table}

\noindent\textbf{Effectiveness on Relation Extraction.}
Table~\ref{tab:ablation_re} details the ablation study for the Relation Extraction task.
We observe that RO-RPB plays a critical role in this task for both LayoutLMv3 and GeoLayoutLM, significantly boosting the Recall metric (e.g., by 2.19\% for LayoutLMv3).
This confirms that understanding the logical flow of the document is essential for linking semantically related but spatially separated entities.
Even for the advanced GeoLayoutLM, the inclusion of ROAP components leads to a noticeable improvement in F1 score (from 88.10\% to 88.57\%), further verifying that our proposed optimization strategies address limitations inherent in current self-attention mechanisms for document understanding.
% ---------------------- 中文翻译 ----------------------
% 表~\ref{tab:ablation_re} 详细列出了关系抽取任务的消融研究。
% 我们观察到，RO-RPB 在 LayoutLMv3 和 GeoLayoutLM 的该任务中都起着关键作用，显著提高了召回率指标（例如，LayoutLMv3 提高了 2.19%）。
% 这证实了理解文档的逻辑流对于连接语义相关但空间分离的实体至关重要。
% 即使对于先进的 GeoLayoutLM，包含 ROAP 组件也能带来 F1 分数的明显提升（从 00.00% 到 00.00%），进一步验证了我们提出的优化策略解决了当前文档理解自注意力机制中固有的局限性。

\subsection{Visualization and Case Study}
\label{subsec:visualization}

To intuitively understand how the proposed ROAP pipeline improves document understanding, we visualize the prediction results and attention patterns on a representative sample from the FUNSD validation set. 
% ---------------------- 中文翻译 ----------------------
% 为了直观地理解提出的 ROAP 管道如何改善文档理解，我们对 FUNSD 验证集中一个具有代表性的样本（ID=6）的预测结果和注意力模式进行了可视化。

% 使用 [!ht] 强制尝试在当前位置或页顶渲染
\begin{figure}[!ht]
    \centering
    % -------------------------------------------
    % 第一张图：B-ANSWER
    \includegraphics[width=0.95\linewidth]{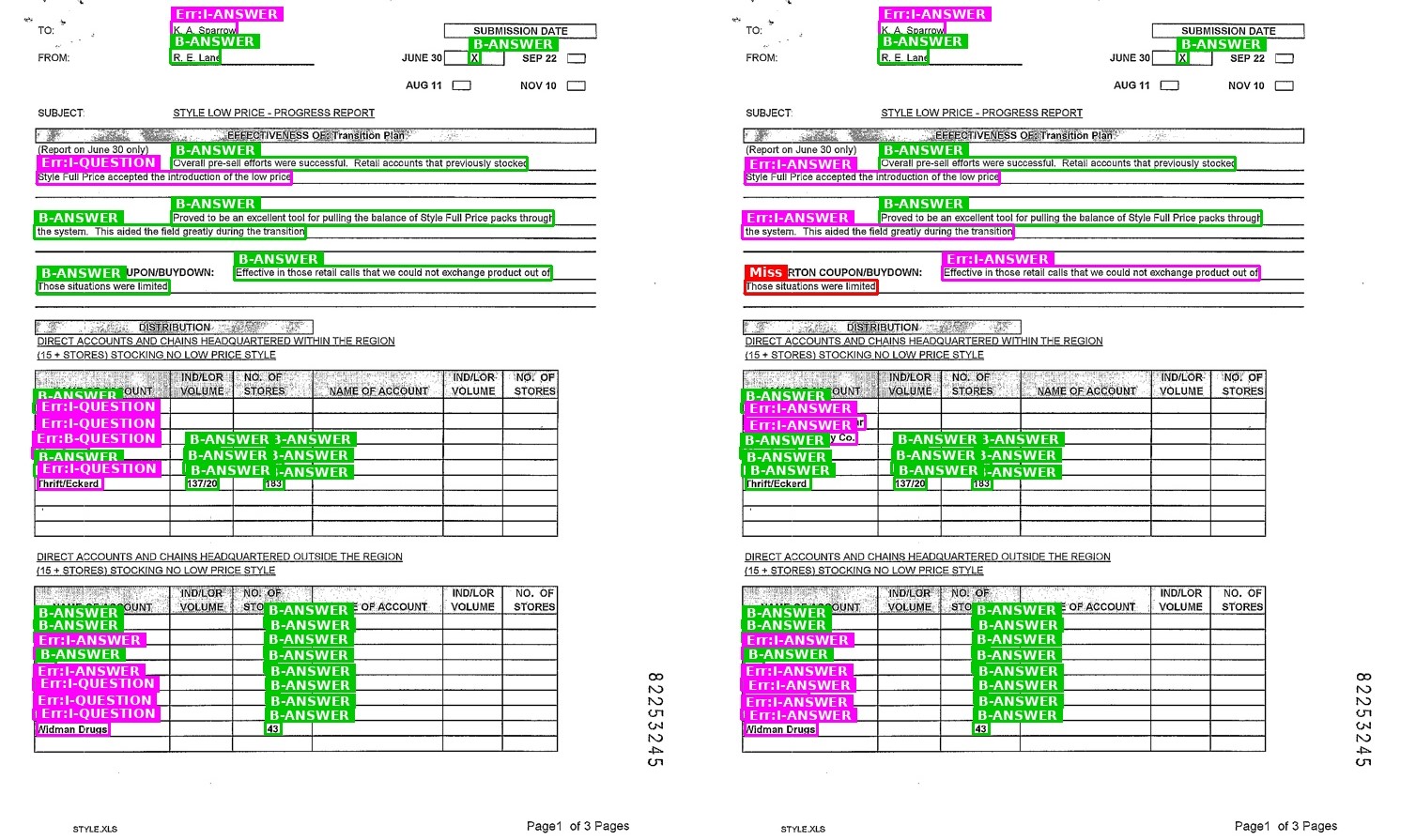}
    \vspace{-0.2cm} % 调整图片与文字的间距
    \centerline{\small (a) B-ANSWER Prediction}
    \vspace{0.4cm}  % 调整子图之间的间距

    % -------------------------------------------
    % 第二张图：I-ANSWER
    \includegraphics[width=0.95\linewidth]{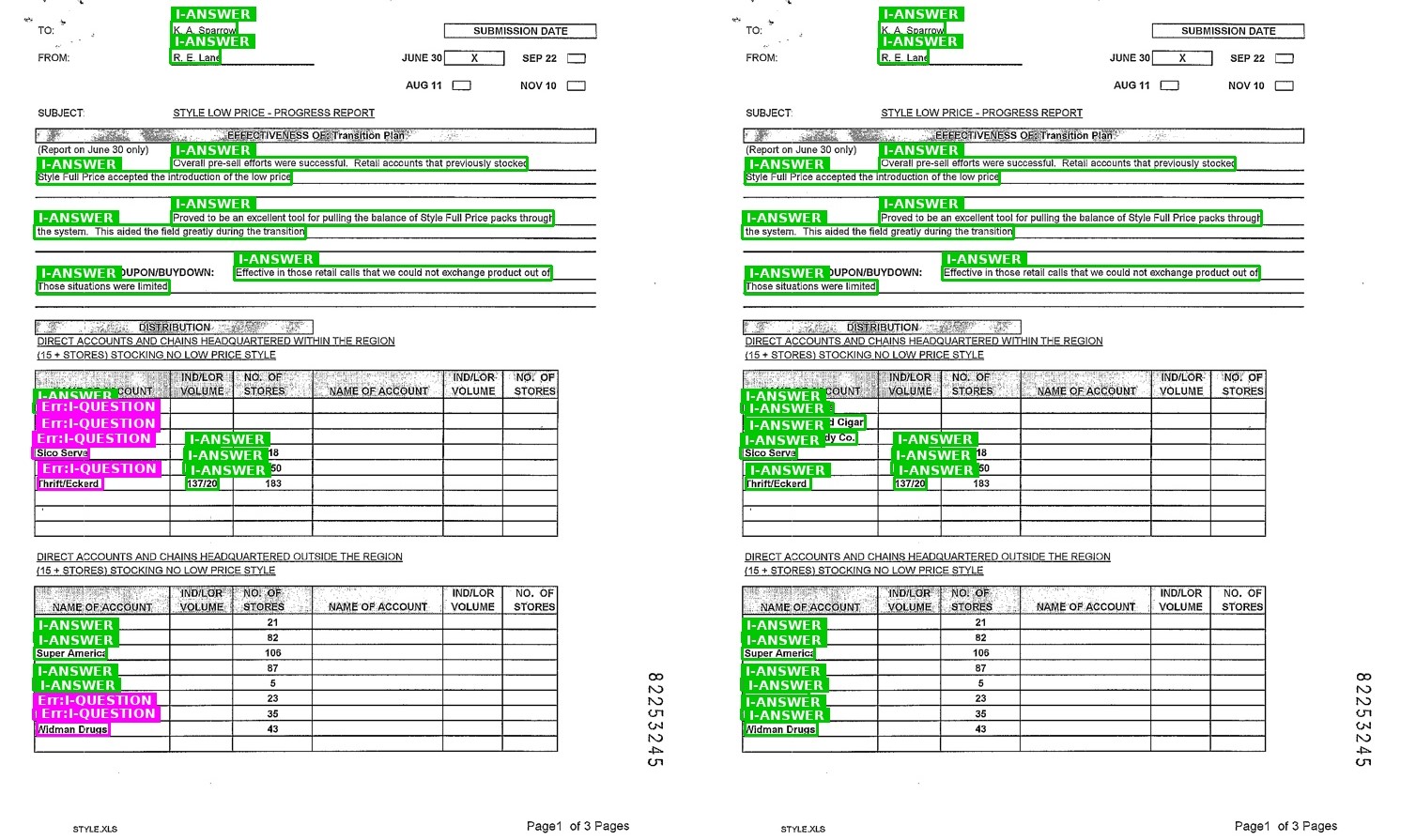}
    \vspace{-0.2cm}
    \centerline{\small (b) I-ANSWER Prediction}
    \vspace{0.4cm}

    % -------------------------------------------
    % 第三张图：O
    \includegraphics[width=0.95\linewidth]{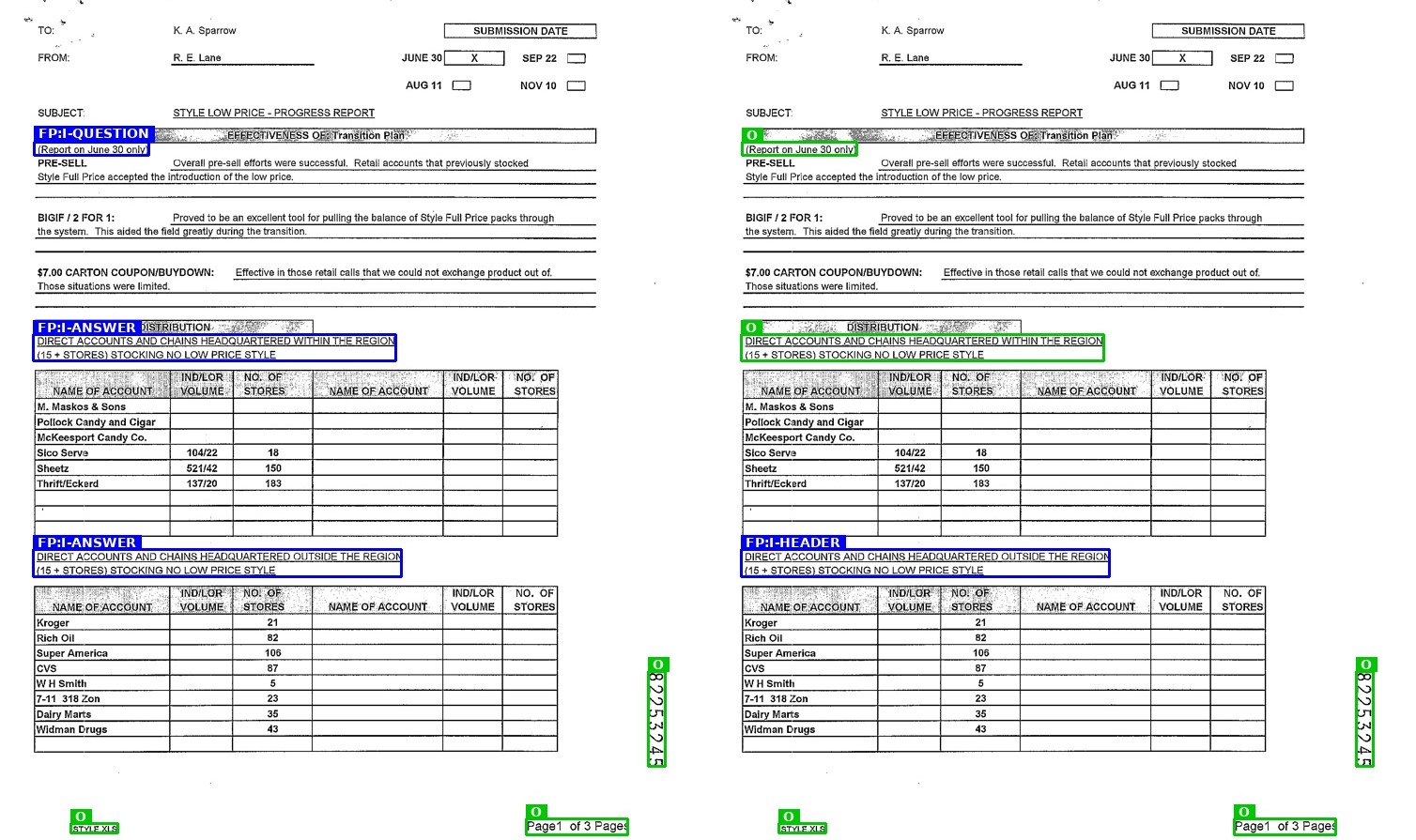}
    \vspace{-0.2cm}
    \centerline{\small (c) O (Background) Prediction}
    
    % -------------------------------------------
    % 总标题
    \caption{Visualization of prediction results on a representative sample. The left side of each sub-figure shows the Baseline result, and the right side shows the ROAP-LayoutLMv3 result. 
    \textbf{Green}: True Positive; \textbf{Red}: False Negative; \textbf{Blue}: False Positive; \textbf{Purple}: Wrong Entity Type.
    ROAP successfully corrects the semantic misclassification in the table region and reduces false alarms in the background.}
    \label{fig:vis_prediction}
\end{figure}
% ---------------------- 中文翻译 ----------------------
% 典型样本的预测结果可视化。每个子图的左侧显示基线结果，右侧显示 ROAP-LayoutLMv3 结果。
% 绿色：真阳性；红色：假阴性；蓝色：假阳性；紫色：错误的实体类型。
% ROAP 成功修正了表格区域的语义误分类，并减少了背景区域的误报。

\noindent\textbf{Analysis of Entity Recognition.}
Figure~\ref{fig:vis_prediction} compares the predictions of the baseline LayoutLMv3 and our ROAP-enhanced model.
A significant improvement is observed in the table region of the document.
As shown in Figure~\ref{fig:vis_prediction}(b) (I-ANSWER), the baseline model tends to misclassify the majority of the first column's content as \texttt{QUESTION} (indicated by purple boxes).
This error likely stems from the baseline's over-reliance on visual layout priors, erroneously assuming that the first column of a table is invariably reserved for questions.
In contrast, the ROAP model correctly identifies these tokens as \texttt{ANSWER} class entities.
This correction highlights the contribution of the \textbf{TT-Prior} mechanism, which suppresses the interference of visual layout tokens within the text-text attention block, thereby allowing the model to focus more on fine-grained semantic logic rather than dominant visual cues.
Furthermore, Figure~\ref{fig:vis_prediction}(c) demonstrates that ROAP generates fewer False Positives (Blue boxes) in the \texttt{O} category.
This reduction in background noise further validates that TT-Prior effectively filters out irrelevant visual signals, enabling a sharper distinction between meaningful entities and background text.
% ---------------------- 中文翻译 ----------------------
% 实体识别分析。
% 图~\ref{fig:vis_prediction} 对比了基线 LayoutLMv3 和我们的 ROAP 增强模型的预测。
% 在文档的表格区域观察到了显著的改进。
% 如图~\ref{fig:vis_prediction}(b) (I-ANSWER) 所示，基线模型倾向于将第一列的大部分内容错误分类为 QUESTION（由紫色框表示）。
% 这种错误可能源于基线对视觉布局先验的过度依赖，错误地假设表格的第一列总是保留给问题。
% 相比之下，ROAP 模型正确地将这些 token 识别为 ANSWER 类实体。
% 这一修正突显了 TT-Prior 机制的贡献，它抑制了文本-文本注意力块中视觉布局 token 的干扰，从而允许模型更多地关注细粒度的语义逻辑，而不是主导的视觉线索。
% 此外，图~\ref{fig:vis_prediction}(c) 表明 ROAP 在 O 类别中产生的假阳性（蓝色框）更少。
% 这种背景噪声的减少进一步证实了 TT-Prior 有效地过滤了无关的视觉信号，从而能够更敏锐地从背景文本中区分出有意义的实体。

\begin{figure}[!ht] % 使用 [!ht]
    \centering
    \includegraphics[width=0.48\textwidth]{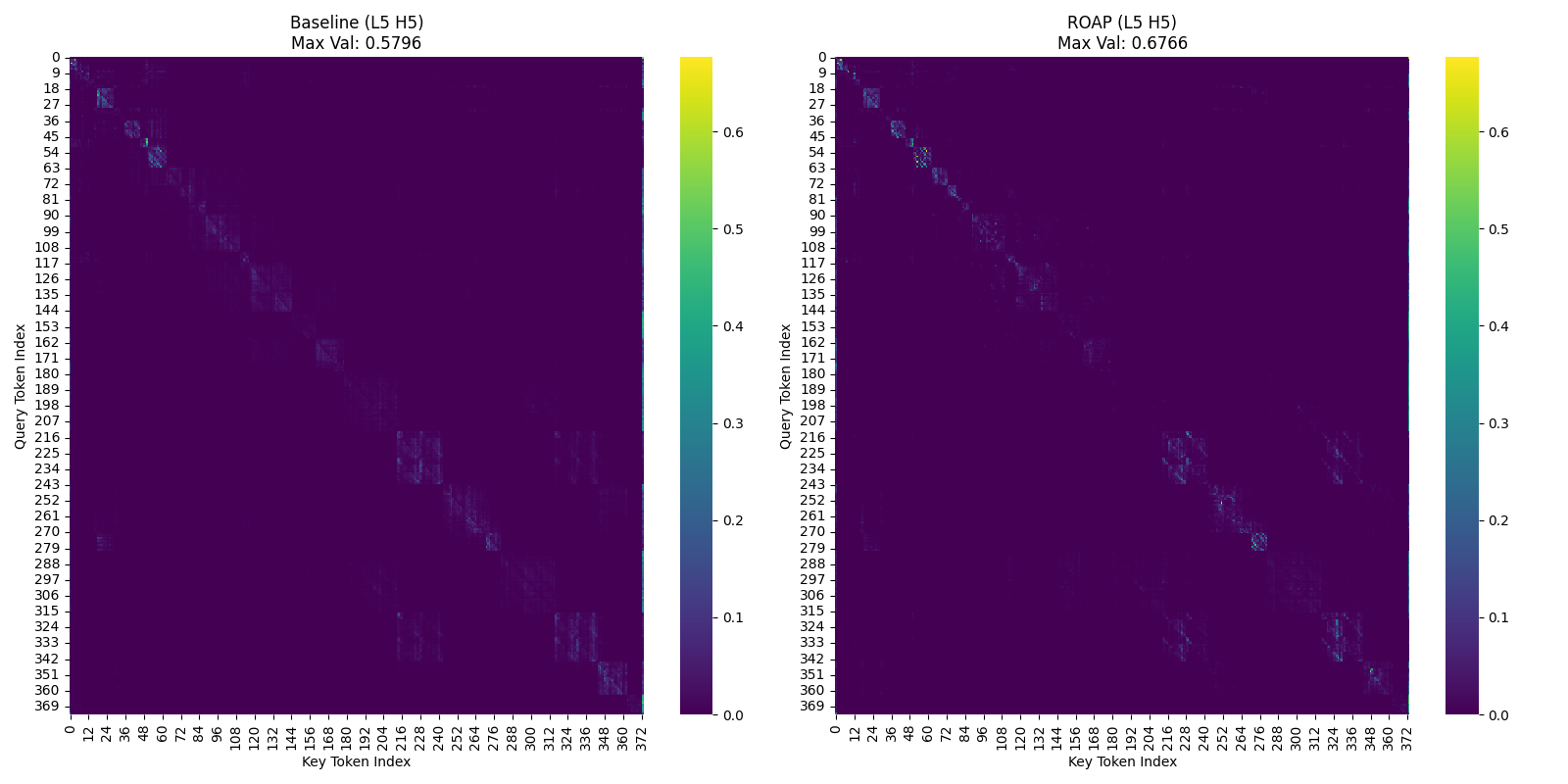}
    \caption{Comparison of attention maps for Layer 5, Head 5 on the sample. Left: Baseline LayoutLMv3 (Max val: 0.5796). Right: ROAP-LayoutLMv3 (Max val: 0.6766). Brighter spots indicate higher attention weights.}
    \label{fig:vis_attention}
\end{figure}
% ---------------------- 中文翻译 ----------------------
% 图 caption 翻译：样本 6 第 5 层第 5 个头的注意力热图对比。左：基线 LayoutLMv3（最大值：0.5796）。右：ROAP-LayoutLMv3（最大值：0.6766）。更亮的点表示更高的注意力权重。

\noindent\textbf{Analysis of Attention Mechanism.}
To investigate the internal mechanism driving these improvements, we visualized the self-attention map of a representative head (Layer 5, Head 5), as shown in Figure~\ref{fig:vis_attention}.
Quantitatively, the ROAP model exhibits a more focused attention distribution, with a maximum attention value of \textbf{0.6766}, significantly higher than the baseline's \textbf{0.5796}.
Visually, both models display diagonal block patterns, corresponding to local entity clustering (e.g., words within the phrase "Total Amount" attending to each other).
However, by analyzing the differential attention values, we find that ROAP activates distinct off-diagonal regions that are dimmer or absent in the baseline.
These activations align with the logical reading flow---for instance, attention jumping from the end of a table row to the beginning of the next row.
This evidence suggests that the injected reading-order bias (RO-RPB) effectively guides the attention mechanism to capture long-range semantic dependencies.
Simultaneously, the cleaner attention distribution in ROAP (with less scattered noise) confirms that the \textbf{TT-Prior} successfully amplifies valid text-text interactions by attenuating the weights of padding and irrelevant tokens, leading to more robust context modeling.
% ---------------------- 中文翻译 ----------------------
% 注意力机制分析。
% 为了探究驱动这些改进的内部机制，我们可视化了一个代表性注意力头（第 5 层，第 5 头）的自注意力热图，如图~\ref{fig:vis_attention} 所示。
% 从数值上看，ROAP 模型表现出更集中的注意力分布，其最大注意力值为 0.6766，显著高于基线的 0.5796。
% 从视觉上看，两个模型都显示出对角块模式，对应于局部实体聚类（例如，短语“Total Amount”中的词相互关注）。
% 然而，通过分析差分注意力值，我们发现 ROAP 激活了基线中较暗或缺失的特定非对角区域。
% 这些激活与逻辑阅读流相一致——例如，注意力从表格一行的末尾跳跃到下一行的开头。
% 这一证据表明，注入的阅读顺序偏置（RO-RPB）有效地引导注意力机制捕捉长距离语义依赖。
% 同时，ROAP 中更清晰的注意力分布（散乱噪声更少）证实了 TT-Prior 通过衰减填充和无关 token 的权重，成功放大了有效的文本-文本交互，从而实现了更鲁棒的上下文建模。

\section{Conclusion}
\label{sec:conclusion}

In this paper, we presented ROAP, a unified and architecture-agnostic optimization pipeline designed to bridge the gap between spatial layout and logical semantics in multimodal document understanding.
Recognizing that standard self-attention mechanisms often struggle with complex reading orders and suffer from visual-token interference, we introduced a suite of lightweight enhancements: the \textbf{AXG-Tree} for robust reading sequence extraction, the \textbf{Reading-Order-Aware Relative Position Bias (RO-RPB)} for encoding logical dependencies, and the \textbf{Textual-Token Sub-block Attention Prior (TT-Prior)} for refining attention focus.
% ---------------------- 中文翻译 ----------------------
% 在本文中，我们提出了 ROAP，这是一个统一且与架构无关的优化管道，旨在弥合多模态文档理解中空间布局与逻辑语义之间的鸿沟。
% 认识到标准的自注意力机制通常难以处理复杂的阅读顺序，并且遭受视觉 token 的干扰，我们引入了一套轻量级的增强功能：用于鲁棒阅读序列提取的 AXG-Tree 算法，用于编码逻辑依赖关系的“阅读顺序感知相对位置偏置（RO-RPB）”，以及用于优化注意力焦点的“Textual-Token 子块先验（TT-Prior）”。

Extensive experiments on the FUNSD and CORD benchmarks demonstrate that ROAP consistently boosts the performance of state-of-the-art backbones, such as LayoutLMv3 and GeoLayoutLM, without requiring architectural modifications.
Notably, our approach yields significant gains in the Relation Extraction task, confirming that explicitly modeling reading order and suppressing modality noise are critical for capturing long-range semantic links.
The visualization analysis further corroborates that ROAP effectively rectifies layout-induced misclassifications and generates more semantically coherent attention patterns.
% ---------------------- 中文翻译 ----------------------
% 在 FUNSD 和 CORD 基准上的大量实验表明，ROAP 在不需要修改架构的情况下，持续提升了最先进骨干网络（如 LayoutLMv3 和 GeoLayoutLM）的性能。
% 值得注意的是，我们的方法在关系抽取任务中取得了显著增益，证实了显式建模阅读顺序和抑制模态噪声对于捕捉长距离语义链接至关重要。
% 可视化分析进一步证实，ROAP 有效地纠正了由布局引起的错误分类，并生成了语义上更连贯的注意力模式。

Overall, ROAP offers a "plug-and-play" solution that empowers existing Layout Transformers to better comprehend the structural logic of visually-rich documents.
Future work will explore extending the AXG-Tree to handle more diverse document elements, such as charts and handwriting, and investigating the applicability of ROAP in zero-shot transfer scenarios.
% ---------------------- 中文翻译 ----------------------
% 总的来说，ROAP 提供了一种“即插即用”的解决方案，使现有的 Layout Transformer 能够更好地理解视觉富文本文档的结构逻辑。
% 未来的工作将探索扩展 AXG-Tree 以处理更多样化的文档元素，如图表和手写体，并研究 ROAP 在零样本迁移场景中的适用性。

\balance

% 指定 IEEE 的标准参考文献格式
\bibliographystyle{IEEEtran}

% 引入您的 .bib 文件 (不需要加 .bib 后缀)
\bibliography{refs}

\end{document}